\documentclass[conference]{IEEEtran}
\usepackage{times}

\usepackage[numbers]{natbib}
\usepackage{multicol}
\usepackage[bookmarks=true]{hyperref}
\usepackage{graphicx}

\usepackage{epsfig} 
\usepackage{mathptmx} 
\usepackage{times} 
\usepackage{amsmath} 
\usepackage{amssymb}  
\usepackage{listings}

\usepackage{scrextend}
\usepackage{xcolor}

\newcommand{\saketh}[1]{{\color{black} #1 \color{black}}}

\begin{document}

\title{NatSGD: A Dataset with \underline{S}peech, \underline{G}estures, and \underline{D}emonstrations for Robot Learning in \underline{Nat}ural Human-Robot Interaction}

\author{Snehesh Shrestha, Yantian Zha, Saketh Banagiri, Ge Gao, Yiannis Aloimonos, Cornelia Fermüller\\ \{snehesh,ytzha,sbngr,gegao,jyaloimo,fermulcm\}@umd.edu \\ University of Maryland, College Park}

\maketitle

\begin{abstract}
Recent advancements in multimodal Human-Robot Interaction (HRI) datasets have highlighted the fusion of speech and gesture, expanding robots' capabilities to absorb explicit and implicit HRI insights. However, existing speech-gesture HRI datasets often focus on elementary tasks, like object pointing and pushing, revealing limitations in scaling to intricate domains and prioritizing human command data over robot behavior records. To bridge these gaps, we introduce \textit{NatSGD}, a multimodal HRI dataset encompassing human commands through speech and gestures that are natural, synchronized with robot behavior demonstrations. \textit{NatSGD} serves as a foundational resource at the intersection of machine learning and HRI research, and we demonstrate its effectiveness in training robots to understand tasks through multimodal human commands, emphasizing the significance of jointly considering speech and gestures. We have released our dataset, simulator, and code to facilitate future research in human-robot interaction system learning; access these resources at \url{https://www.snehesh.com/natsgd/}


\end{abstract}

\IEEEpeerreviewmaketitle

\section{Introduction}
Human communication often involves the concurrent use of language and gestures, as noted by Tomasello (\textit{p}.230 in \cite{tomasello2010origins}). Language-based communication excels at conveying explicit knowledge, while behavior-based communication excels at expressing tacit knowledge. However, for tasks that encompass both explicit and implicit aspects, such as cooking and cleaning, a hybrid approach becomes essential.

\begin{figure*}[ht]
\centering
\includegraphics[width=\textwidth]{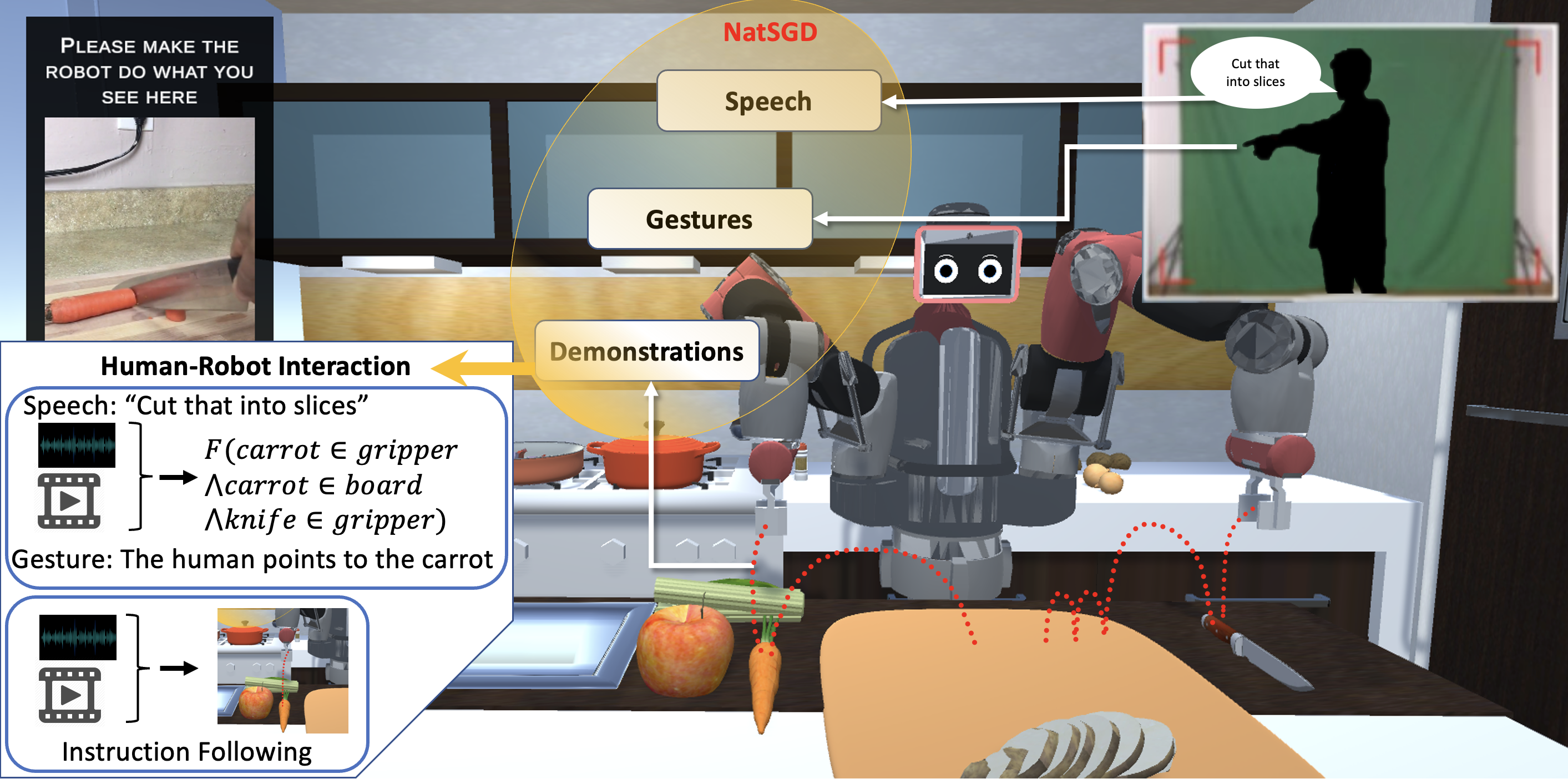}
\caption{NatSGD contains speech, gestures, and demonstration trajectories for  everyday food preparing, cooking, and cleaning tasks. The \textit{NatSGD} dataset potentially enables the learning of complex human-robot interaction tasks due to the rich interaction modalities and strong supervising signals at both trajectory-level (demonstrations) and symbolic-level (ground-truth activities that match humans' intention).}
\label{fig:intro}
\end{figure*}

The integration of robots into human collaboration, emulating a more natural interaction, could notably alleviate cognitive burdens on humans. This underscores the necessity for datasets facilitating the adept assimilation of natural human guidance, encompassing both language and gestures, by robots. However, a notable trend in the majority of Human-Robot Interaction (HRI) datasets, including those for prominent collaborative robots like Google Home, Amazon Alexa, and Apple Siri, is their exclusive reliance on speech as the communication medium \cite{novoa2017multichannel,james2018open,vasudevan2018object,narayan2019collaborative,padmakumar2022teach}. Conversely, a parallel stream of HRI datasets, such as \cite{pisharady2015recent,shukla2016multi,mazhar2018towards,chen2018wristcam,chang2019improved,gomez2019caddy,neto2019gesture,nuzzi2021hands,de2021introducing}, focuses solely on gestures during human-robot interactions \cite{luan2016reliable}. A subset of works does introduce HRI datasets combining speech and gestures \cite{matuszek2014learning,rodomagoulakis2016multimedia,azagra2017multimodal,chen2022real}. However, these efforts predominantly center on perception-oriented tasks, such as object recognition based on simple colors and shapes, or tasks like object manipulation. This contrasts with real-world complexities, where individuals employ a myriad of features, vocabulary, and styles to denote objects and actions.

Consequently, our focus lies in creating a pioneering Human-Robot Interaction (HRI) dataset that satisfies the following objectives: 1) Emulate natural human communications encompassing both speech and gestures; 2) Equip robots with the capability to comprehend intricate tasks, such as cooking and cleaning, which hold significance in our daily lives; and 3) Integrate demonstration trajectories, necessitated by the multifaceted nature of the tasks. To realize this, we devised a Wizard of Oz (WoZ) experiment \cite{Dahlback1993-el}, wherein participants interacted organically with what they believed to be a highly competent autonomous humanoid. Leveraging the insights from these experiments, our primary contribution materialized in the form of \textit{NatSGD} -- a dataset for training robots to interact with humans in a natural way (Fig. \ref{fig:intro}) that introduces previously unexplored speech and gesture data, complemented by paired robot demonstrations that lead to real-world-like tasks' fulfillments. 

As far as our knowledge extends, the \textit{NatSGD} dataset is the first to encompass speech, gestures, and demonstration trajectories, which equips robots to learn intricate tasks like cooking and cleaning -- activities deeply embedded in our daily routines. Consequently, the \textit{NatSGD} dataset presents a wealth of research prospects, facilitating the training of robots for seamless and natural interactions with humans. A pivotal HRI challenge is the comprehension of tasks specified through diverse modalities, encompassing speech and gestures. We term these tasks as Multi-Modal Human Task Understanding. Addressing this challenge entails tackling two key issues: 1) The formulation of task representations that can delineate intricate relationships between task subcomponents; and 2) The development of a mapping mechanism to translate multi-modal commands into these task representations. Notably, the latter issue remains an unexplored frontier, involving the transformation of language and gesture inputs into task features -- an endeavor that warrants further exploration.

In our research, we present three key contributions: Firstly, the introduction of the \textit{NatSGD} Dataset, which comprises 1143 commands issued by 18 individuals, covering 11 actions, 20 objects, and 16 states, with each command accompanied by an expert demonstration. Secondly, we demonstrate the enhanced performance achieved when comprehending human-specified tasks through the combined use of speech and gestures, surpassing the results obtained using either modality in isolation. Finally, we have developed a photorealistic simulated environment integrated with a Robot Operating System (ROS) interface, which greatly streamlines both Reinforcement Learning and Sim2Real \cite{hofer2021sim2real} processes.

The remainder of this document is structured as follows: Following an examination of relevant prior research, we will proceed to provide an in-depth exploration of the \textit{NatSGD} dataset collection process. 



\section{Related Work}

As mentioned in the Introduction section, there are various datasets that adopt either speech or gestures as communication modalities. In this section, we focus on discussing the datasets that involve both communication modalities. We further discuss how those datasets can enable the learning of HRI systems.

\subsubsection{Datasets for Speech-and-gesture-based HRI}
Many Speech-and-Gesture datasets do not involve a robot. For example, the HRI datasets presented in \cite{azagra2017multimodal} and \cite{rodomagoulakis2016multimedia} completely focus on perceptual tasks. \cite{azagra2017multimodal} aims at solving tasks like interaction recognition, target object detection, and object model learning, whereas \cite{rodomagoulakis2016multimedia} addresses tasks like visual gestures recognition and audio command phrases recognition. \cite{lee2019talking} and \cite{kucherenko2021large} presents gesture-speech multimodal human-human conversation datasets that have a potential of training robots for HRI scenarios. Unlike the above datasets, there are also a few that demonstrate values for robotics research. \cite{matuszek2014learning} introduces a HRI dataset in which humans use both gesture and language to refer to an object. Robots learn to recognize the referred object and push that object away. Similarly, \cite{chen2022real} collects a speech and gesture dataset and demonstrates its value on a robotic pick-and-place task in HRI settings. The pick-and-place system works by having robots recognize target locations by observing humans' speeches and gestures and then inputting the target region to a motion planner. In contrast to these datasets, our \textit{NatSGD} dataset includes demonstration trajectories and thus enables the robot learning of more complex tasks like cooking and cleaning. The importance of including demonstration trajectories in HRI datasets is also acknowledged by a recent dataset paper \cite{padmakumar2022teach}, which however does not consider gestures.

\subsubsection{Robot Applications on HRI Datasets}
The HRI community also lacks robot learning works that involve training deep neural networks on HRI datasets. The aforementioned HRI works \cite{matuszek2014learning,chen2022real} put more focus on the perceptual aspects of using their collected datasets and therefore only applies perception outputs (e.g. the pose of recognized object) to an engineered robotic system. Similarly, \cite{krishnaswamy2020formal} provides a formal analysis on the multimodal recognition of objects referred in HRI. \cite{ahuja2020no,habibie2021learning,kucherenko2021multimodal,habibie2022motion} research the correlation between speech and gestures that could lead to better gesture recognition or generation functions. However, these works do not really involve robots. In contrast to previous studies, we assess our dataset by training robots to comprehend tasks specified by humans using both natural language and gesture commands.

\section{\textit{NatSGD} Dataset}

The \textit{NatSGD} dataset is designed to fill critical gaps in existing datasets, with a focus on achieving naturalness and practicality, bridging the gap between research and real-life applications in human-robot interaction. \saketh{The \textit{NatSGD} dataset encompasses tasks and scenarios from real-world kitchen settings, including food preparation, serving, and cleanup activities (elaborated in Appendix \ref{appen:storyline}).} 

To be truly valuable for robot learning, a dataset must be comprehensive and representative. On the human side, meticulous control is exercised over participants and their behaviors to mitigate bias. On the robot side, the system must be capable, and the simulator must exude realism to instill trust and belief in human users. The interactions should unfold naturally, driven by participants' own volition, free from external priming. Furthermore, the dataset's versatility is key, making it applicable to a wide range of robot learning tasks. This section outlines the distinctive aspects of \textit{NatSGD} that render it innovative and practical.

\begin{figure}[ht]
\centering
\includegraphics[width=\columnwidth]{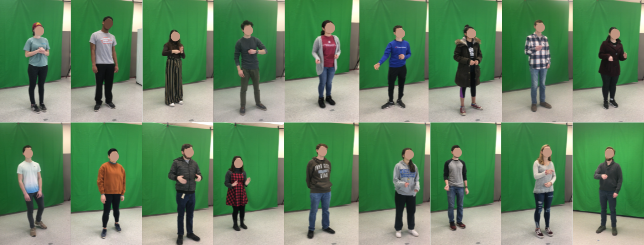}
\caption{In this example, the participants are commanding the robot to cut onions where the participant's natural choice of communication is diverse.}
\label{fig:all_participants_cut_onion_diverse}
\end{figure}

\subsection{Humans and Bias}
Ensuring fairness and a deep understanding of the dataset is paramount. Individuals harbor implicit and explicit biases, which can permeate the data. It's crucial to address biases both from a technical machine learning perspective and from a societal standpoint. Factors such as gender, age, expertise, culture, and personality are considered in participant selection. Additionally, individual experiences, including workload assessments using NASA-TLX \cite{hart2006nasa}, and participants' perceptions of the robot are recorded. The design of the robot's face, name, and movements takes into account its perceived gender, approachability, and naturalness, as elaborated in Appendix \ref{appen:robot_face_and_name}.

In this experiment, a total of eighteen participants, aged between 18 and 31 years (with a mean age of 20.91\textpm3.75), took part, evenly divided between genders (9 male and 9 female). Personality traits were assessed, resulting in scores for Extroversion (5.56\textpm2.09), Agreeableness (9.17\textpm1.15), Conscientiousness (8.17\textpm1.38), Emotional Stability (7.56\textpm1.79), and Openness to Experience (7.89\textpm1.47). To ensure diversity, an equitable distribution of participants from technical and non-technical backgrounds was selected based on their familiarity with high-tech games and toys, as well as their educational or professional backgrounds in computer science and engineering. Notably, none of the participants had prior experience interacting with robots before participating in this study. For more comprehensive participant details, please refer to Appendix. \ref{appen:participants}.

The human speech was transcribed into text, and we utilized the Glove model \cite{pennington2014glove} and BART \cite{lewis2019bart} to extract speech embeddings. Similarly, human poses were extracted using OpenPose \cite{james2018open}, representing a sequence of human pose vectors as the gesture feature.

\subsection{Robot and Realism}
Everyday activities like cutting vegetables, which are simple for humans, pose significant difficulties for robots. Conducting Human-Robot Interaction (HRI) experiments with real robots, especially in real-time settings for collecting natural interaction data, is exceptionally demanding. Drawing inspiration from recent advancements in sim2real research \cite{abeyruwan2022sim2real, kaspar2020sim2real}, we designed and developed our photorealistic simulator. Leveraging Unity3D \cite{Unity_Technologies_undated-qy}, we created a real-time robotics simulator that researchers can swiftly control based on participants' commands. This simulator not only opens avenues for advancing HRI research in complex real-world environments and tasks but also paves the way for future sim2real research in HRI contexts.

\begin{figure*}[ht]
\centering
\includegraphics[width=\textwidth]{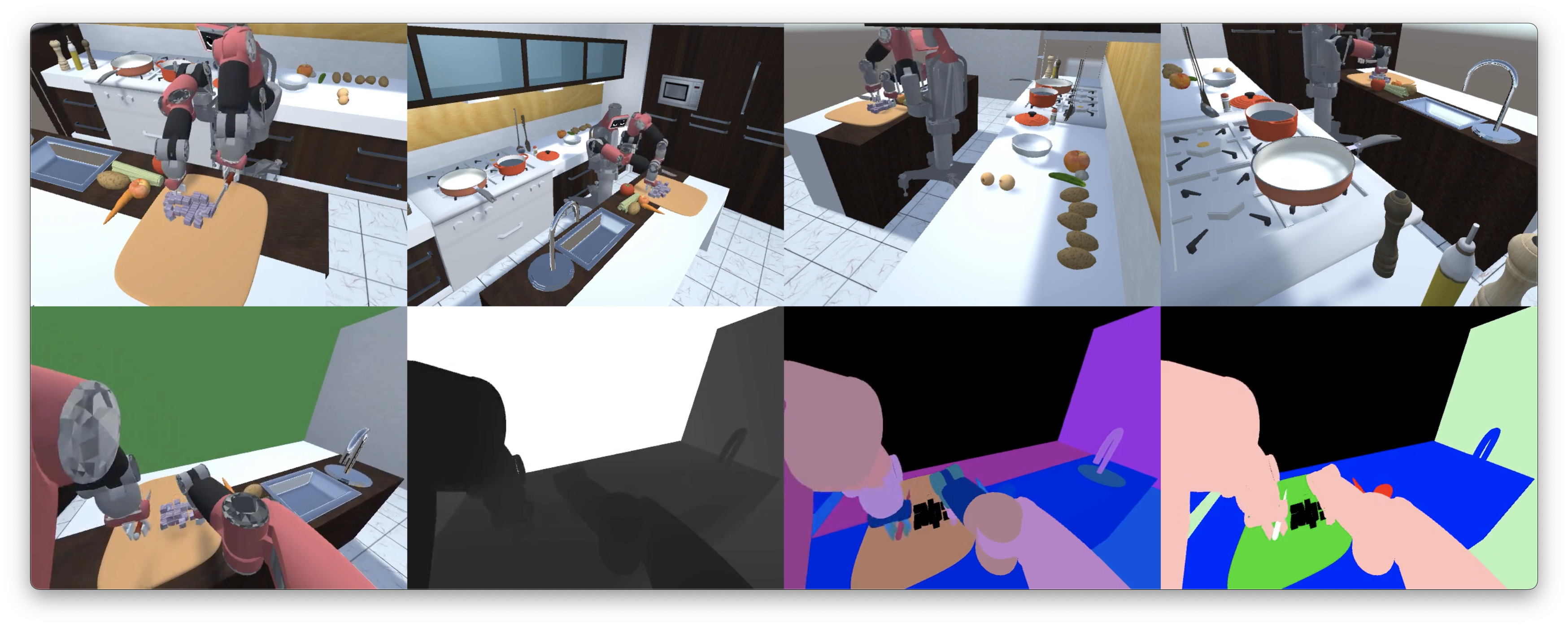}

\caption{Illustration of Our Simulator: Baxter Slicing Onions. This simulator offers a range of multi-view perspectives, capturing both the environment and the robot through static and mobile camera angles. The top row presents the human-first-person view, along with overhead (top left), and kitchen counter (bottom right and bottom left) camera angles. The bottom row showcases the robot's egocentric viewpoint, encompassing RGB, depth, distinctive object segmentation, and category-based semantic segmentation. These diverse perspectives empower the robot to learn and execute tasks based on human speech-gesture commands, as well as its autonomous assessment of the surroundings and object conditions.}
\label{fig:natsgd_multi_cam_views}
\end{figure*}
\begin{figure}[ht]
\centering
\includegraphics[width=\columnwidth]{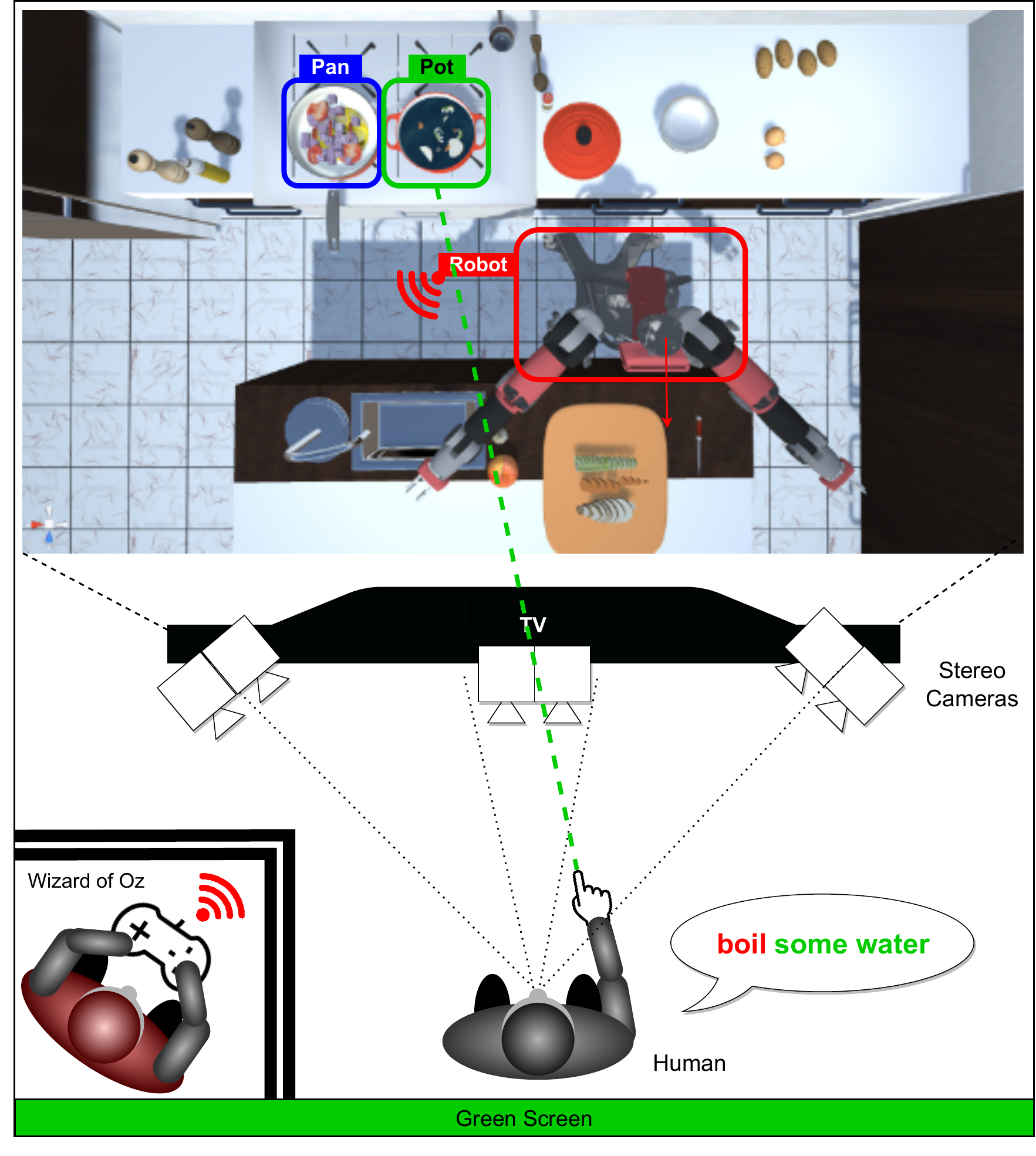}
\caption{NatSGD experiment setup top view. In this figure, the participant is  pointing at the pot on the right stove and asks the robot to boil some water implying the right burner needing to be turned on. On the bottom left shows the location of the Wizard of Oz hidden to the participant making observations and controlling the robot.}
\label{fig:NatComm_Dataset_TopAndFrontView_PourVeges}
\end{figure}

For a deeper dive into our simulator (illustrated in Fig. \ref{fig:natsgd_multi_cam_views}), the \textit{NatSGD} Robot Simulator was constructed using Unity 3D in conjunction with a customized ROS plugin \cite{Bischoff_rossharp}. The system operated on a computer equipped with an Intel i7 processor and 16GB of RAM, connected to a 55" TV to ensure smooth real-time rendering and processing. In the top right corner of the screen, a live camera feed of the participant was overlaid (as depicted in Fig. \ref{fig:intro}), providing participants with the perception that the robot could see them and helping them stay within the frame, as illustrated in Fig. \ref{fig:NatComm_Dataset_TopAndFrontView_PourVeges}. Real-time inverse kinematics (IK) for the robot's head movement and arms were implemented using BioIK \cite{Starke2017-cx, Starke2018-pb}. The robot would maintain eye contact with target objects during tasks, demonstrating its attention, and then return its gaze to the participant when ready for the next interaction. The simulator's detailed output encompasses perspectives from both the human and the robot, object states, and robot trajectories, as extensively described in Appendix \ref{sec:ground_truth_labels}. Notably, our simulator excels at handling activities that involve causal sub-activities, such as pouring and cutting. When the robot undertakes the task of cutting a tomato, it must first locate a knife, secure it with a grasp, approach the tomato on the chopping board, maintain stability, execute the initial cut, and then proceed with successive slices—a level of complexity that our simulator adeptly replicates.

\subsection{Interaction and Naturality}\label{sec:Interaction_and Naturality}

Capturing authentic human behavior is paramount for effective human-robot interaction. Natural cues, characterized by their unstructured, multimodal nature and implicit contextual components, offer invaluable insights for robot learning. These cues often comprise contradictory phrases and mechanisms for correction. However, inducing genuinely natural human behavior within a controlled laboratory setting is a formidable challenge. Drawing inspiration from prior research and conducting a series of pilot studies, we devised a Wizard-of-Oz (WoZ) experiment design \cite{Dahlback1993-el}. This approach was instrumental in eliciting spontaneous, natural human behavior for our research endeavors.

\subsubsection{Pilot Studies}

In our pursuit of achieving the ideal balance between real-world spontaneity and controlled laboratory conditions, our Wizard-of-Oz (WoZ) experiment design involves subtly deceiving participants into perceiving the robot as fully autonomous. To ensure the effectiveness of this approach, we conducted a series of pilot studies aimed at validating various factors that could influence participant behavior. These studies served to validate both independent and dependent control variables, as well as to refine our workflow. Specifically, we explored the impact of:
\begin{itemize}
  \item \textbf{Background Noise:} Investigating the role of background noise (detailed in the Background Noise section in the appendix).
  \item \textbf{Perceived Robot Personality and Capability:} Evaluating how participants' perceptions of the robot's personality and capabilities, as influenced by the robot's face and name (as outlined in Appendix. \ref{appen:robot_face_and_name}), affected their interactions.
  \item \textbf{Staging:} Experimenting with strategies to keep participants engaged throughout the study (as discussed in Section \ref{sec:staging}).
  \item \textbf{Priming Effects from Practice Sessions:} Assessing the potential influence of priming effects resulting from practice sessions (as described in Appendix. \ref{appen:practice_session}).
  \item \textbf{WoZ Clues:} Exploring any WoZ-related cues that participants might use to discern hidden motives (as documented in Appendix. \ref{appen:woz_cues}).
  \item \textbf{Experiment Instructions:} Analyzing the effects of the instructions provided to participants (covered in the Instruction section of Appendix. \ref{appen:instructions_human}).
\end{itemize}

The insights gained from these pilot studies played a pivotal role in shaping our final experiment design decisions.

\subsubsection{Expert Demonstration and WoZ Control Policy}
Throughout both practice and data collection sessions, the researcher facilitator, acting as the ``wizard," orchestrates the workflow, advancing it as long as the participant's commands pertain to the ongoing task and are intelligible. When the robot comprehends a command, it signals its understanding with a nod (head motion up and down). However, in the case of unintelligible or unrelated commands, the robot responds with a confused facial expression. For instance, if a participant instructs the robot to move right, the wizard will make the robot execute the corresponding action. Yet, if the participant's speech is unclear, leading to difficulty in comprehension, the wizard prompts with a puzzled robot expression, seeking clarification from the participant. This repair prompt, deliberately designed to be somewhat ambiguous, encourages participants to experiment with different instruction strategies for the same task. Each participant is afforded a maximum of five attempts per command. If unsuccessful after these attempts, or in rare cases where the Unity game freezes or Baxter encounters difficulties in grasping objects for over 30 seconds, the task may be skipped.

\subsubsection{Staging Techniques \cite{thomas1995illusion}}
\label{sec:staging}
As the participant and the robot do not share the same physical space, one challenge that we observed was when the robot was not directly facing the person could lead to break in flow, context, visibility, and frame of reference. To overcome this, we borrow techniques from the 12 principle of animation, specifically \textit{staging} \cite{thomas1995illusion} to gently pan-and-rotate the camera to a camera pose that gives a clear view of the key event. This is akin to participant walking over and standing near the robot where the task is being done. For example, if the robot is pouring oil into the pan, we pick camera angle such that the pan is in the center, with the robot in the background, and oil visible to one side of the screen as shown in Fig \ref{fig:natsgd_multi_cam_views} top right image.

\subsection{Versatile Applicability}
\label{sec:versatile_applicability}
We believe the dataset to be more useful if it can be applied towards multiple use cases. To this end, \textit{NatSGD} consists of data in multiple modalities and multiple vantage points from multiple subjects. It consists of long continuous sequences with versatile ground truth labels that are annotated by multiple human annotators or computationally generated. The ground truth labels include object labels, bounding boxes, event labels, logic formula representation of tasks and subtasks, and human commands (speech and gestures), and demonstration trajectories (robot trajectories generated from WoZ teleportation). Based on \textit{modality}, each human command was annotated for consisting relevant \textit{speech} and \textit{gesture}. We then further distinguished them based if they \textit{referred} to \textit{objects} or \textit{action}. The gesture is annotated as containing task-related (\textit{intentional}) and task-unrelated (\textit{unintentional}) gestures. Finally, for all gestures, we also annotated the role of each \textit{body part} in task specific action/ object reference and non-intent movements as illustrated in Fig. \ref{fig:natcom_dataset_tree}. Please refer to section \ref{sec:dataset_structure} for dataset structure and Appendix. \ref{sec:ground_truth_labels} for more details.

Therefore, \textit{NatSGD} holds promise for a wide range of applications, including low-level tasks like gesture recognition, speech recognition, and object detection. Particularly beneficial for semantic-level gesture recognition, it pairs gestures like pointing with corresponding speech cues for tasks such as ``cut it into pieces." Moreover, \textit{NatSGD}'s utility extends to classifying gesture properties, differentiating between intentional and unintentional gestures, which has practical implications in Human-Robot Interaction (HRI). At a higher level, the dataset empowers robot learning tasks, exemplified by multi-modal human task understanding, where comprehending tasks from both speech and gestures is crucial. This high-level task serves as a benchmark for the dataset and is elaborated in the Sec. \ref{sec:multi-modal-human-task-understanding}. With its diverse potential, \textit{NatSGD} opens up an array of research possibilities for the broader community to explore.

\subsection{Reliable and Robust Data}
\label{sec:data_reliability}
We carefully take iterative steps to prepare the data to ensure integrity, quality, validity, and fairness. We calibrated and synchronized all the cameras audio-video. We compressed the final video to a lossless high quality data format. Multiple people annotated our data using Fast Event Video Annotation Tool (FEVA \cite{shrestha2023feva}) and FEVA crowd (see Fig. \ref{fig:feva_p40} and  Fig. \ref{fig:feva_p61} in Appendix). \saketh{When annotating Linear Temporal Logic (LTL), we ensured that the human annotators were experts in LTL. They carefully analyzed every video segment, speech, and previously completed subtask before crafting the ground-truth LTL formula. Each label received a minimum of two annotations, followed by up to three rounds of consensus verification to improve both efficiency and reliability.}

\saketh{

\subsection{Dataset Structure}
\label{sec:dataset_structure}

The dataset is configured as follows: Each command instance is assigned a unique identifier, accompanied by tags for the \textit{participant}, \textit{video}, \textit{speech}, \textit{gesture}, \textit{LTL formula}, \textit{robot trajectories}, and \textit{state}. 

Additionally, it includes the timing for the \textit{onset} and \textit{offset} of both \textit{speech} and \textit{gesture}, along with the actual \textit{speech} text. 

Tasks are organized into groups according to the \textit{action-object} pair, with additional details provided below. The structure includes one-hot encoding for \textit{speech}, effectively connecting \textit{speech} elements to particular \textit{objects} and \textit{actions}. Similarly, one-hot encoding for \textit{gesture} is employed, establishing links between \textit{gestures} and their associated \textit{objects} and \textit{actions}.

Gestures are classified into those related to the \textit{task} and those without intent, with one-hot encodings for all six body parts involved in \textit{task action gestures}, \textit{task object gestures}, and \textit{non-intent gestures}.

\subsubsection{Action Groups}
Eleven distinct action groups are featured, including: \textit{Add, Clean Up, Cut, Fetch, Put On, Serve, Stir, Take Off, Transfer, Turn Off, and Turn On.}


\subsubsection{Interaction Object Groups}
Objects are categorized into twenty groups, which are:
\begin{itemize}
    \item Food Ingredients: \textit{Carrot, Celery, Pepper, Potato, Salt, Soup, Spices, Tomato, Veges}
    \item Cooking Utensils and Tools: \textit{Cutting Board, Knife, Bowl, Spatula, Ladle.}
    \item Cookware: \textit{Lid, Oil, Pan, Pot.}
    \item Kitchen Appliances and Fixtures: \textit{Stove, Sink.}
\end{itemize}

\subsubsection{Object States and Attributes}
Object states and attributes include:
\begin{itemize}
    \item States: \textit{Whole, Cut (Pieces), On, In, Out, Covered, Uncovered, Turned On, Turned Off, and Contains.}
    \item Location and Rotation: Annotated with six degrees of freedom for location and rotation.
    \item Participant Interaction: Gaze and pointing gestures are categorized into \textit{Left, Middle, Right directions.}
\end{itemize}

\subsubsection{Robot States}
The robot state comprises both global position and orientation, alongside  each arm's seven joints' rotation angles.


\subsubsection{Additional Dataset Information}
\begin{itemize}
    \item Annotations include IDs for the participant, video, speech, gesture, pose, and state, alongside the time onset and offset for speech and gestures, LTL formula, and detailed text for speech.
    \item Data Formats: Available Python dictionary compressed numpy file formats, with Python scripts for loading and parsing.
    \item External References: Video ID, pose ID, and state ID reference external files in their corresponding folders.
\end{itemize}

}

\section{Multi-Modal Human Task Understanding}\label{sec:multi-modal-human-task-understanding}
The task of comprehending tasks articulated by humans is of paramount importance for achieving seamless human-robot interaction. However, the challenge of understanding such tasks through speech and gestures within the \textit{NatSGD} dataset is far from straightforward. Given the dataset's focus on capturing natural human communication, where interactions unfold as if between humans, establishing a direct one-to-one correspondence between a speech-gesture pair and a singular activity is elusive. To elucidate, consider the statement "pour soup into the bowl." This seemingly simple instruction encapsulates a cascade of conditional sub-tasks: procuring the bowl, positioning it in proximity to the pot, accounting for the pot's lid, uncovering it if needed, identifying and retrieving the ladle, scooping soup, moving the ladle while avoiding spillage, and ultimately pouring the content into the bowl. This iterative process continues until a desired quantity is achieved. This intricate interplay underscores the necessity for an enhanced task representation.

In light of this, we propose to address this complexity by harnessing a mapping technique that translates speech and gestures into a Linear Temporal Logic (LTL) formula, which serves as a task descriptor \cite{pnueli1977temporal,kesten1998algorithmic,konur2013survey}.

\subsection{Linear Temporal Logics}
\label{sec:LTL}
Linear Temporal Logic (LTL) is a modal logic system capable of expressing temporal relationships among events in the form of formulae and performing logical reasoning based on those formulae. LTL has found practical utility in robotics and planning research, facilitating the formulation of high-level reactive task specifications \cite{finucane2010ltlmop,guo2013revising,baran2021ros}. An illustrative instance is the conversion of a natural language task directive like "Go to all rooms on the first floor and then go to the second floor" into an LTL formula: "$F (\forall \textnormal{room} \in \textnormal{rooms} \textsf{ s.t. } \textnormal{floor}_1 \bigwedge X F(\textnormal{floor}_2))^1$". In this formula, the operators $F$ and $X$ signify "finally" and "next," respectively. The $F$ operator mandates that all variables contained within it must hold true to satisfy the formula. However, the process of converting language to LTL becomes intricate in the context of natural human-robot interactions. To achieve an accurate inference of the task specified by a human, a robot must also consider human body language and its own current state.

For instance, consider the speech "Cut that into slices," as depicted in Fig. \ref{fig:intro}. To interpret the referent "that," the robot must leverage the accompanying gesture of the human pointing to the carrot. Additionally, even though the human did not explicitly mention the knife, the command implies that the robot should grasp the knife. Consequently, the LTL formula derived in this scenario becomes:

\begin{addmargin}[2em]{0em} 
\begin{scriptsize}
\texttt{X (G (C\_Carrot U Carrot\_FarFrom\_CT) \& G F (Carrot\_OnTopOf\_CB \& Carrot\_CloseTo\_CB) \& X ( G (C\_Knife U Knife) \& G (C\_Knife U Knife\_FarFrom\_CT) \& G (C\_Knife U Knife\_OnTopOf\_Carrot) \& G ((C\_Carrot \& Knife\_CloseTo\_Carrot) U Carrot\_Pieces)))}
\end{scriptsize}
\end{addmargin}

Here, ``X," ``F", ``U", and ``G" are LTL operators representing ``neXt", ``Finally", ``Until", and ``Globally (Always)" respectively. Additionally, ``C\_Carrot", ``Carrot\_FarFrom\_CT", ``Carrot\_CloseTo\_CB", and ``Carrot\_Pieces" are grounded predicates signifying relations such as ``gripper is close to the carrot", ``carrot is far from the counter top", ``carrot is close to the cutting board", and ``carrot state where it has been cut into pieces" respectively.

\subsection{Problem Formulation}

We present the problem of speech-gesture-conditioned task understanding as a translation task that converts a pair of speech and gestures into an LTL formula. This can be seen as a variant of the multi-modal machine translation problem.

In this formulation, the inputs consist of the following components:

\begin{enumerate}
  \item Source Language Vocabulary $\mathcal{V}_{\text{source}}$: This set encompasses words in the source language $\mathcal{L}_{\text{source}}$, which serves as the language in which human instructions are provided.
  \item Gesture Contexts $\mathcal{Z}$: These sequences are derived from human skeleton data and encapsulate diverse gestures made during interactions.
  \item Target LTL Vocabulary $\mathcal{V}_{\text{target}}$: This set encompasses LTL symbols in the target language $\mathcal{L}_{\text{target}}$, which is used to formulate LTL task descriptions.
\end{enumerate}

To solve the speech-gesture-conditioned task understanding problem, the process involves taking a Source Text Sequence $\mathbf{X} = (x_1, x_2, \ldots, x_n)$. This sequence comprises $n$ words from the source language vocabulary, where each word $x_i$ is drawn from $\mathcal{V}_{\text{source}}$. Additionally, there's the Gesture Context $\mathbf{z} = (z_1, z_2, \ldots, z_t)$, which consists of a sequence of $t$ steps of human skeleton data. Each step $z_i$ is a representation of the skeleton's state at that moment, and $\mathbf{z}$ belongs to the set $\mathcal{Z}$.

The desired outcome is a matching target LTL formula $\mathbf{Y} = (y_1, y_2, \ldots, y_m)$. Each element $y_i$ in this sequence belongs to the target LTL vocabulary $\mathcal{V}{\text{target}}$. The solution must ensure that $\mathbf{Y}$ effectively conveys the meaning of the input $\mathbf{X}$ within the context of the target language $\mathcal{L}{\text{target}}$. This task is further complicated by the presence of linguistic and cultural distinctions between the source language $\mathcal{L}{\text{source}}$ and the target language $\mathcal{L}{\text{target}}$.

The learning objective is to minimize the distance between a predicted LTL formula $\hat{Y}=(\hat{y}_1, \hat{y}_2, \ldots, \hat{y}_m)$ and the corresponding target LTL formula $\mathbf{Y} = (y_1, y_2, \ldots, y_m)$:

\begin{equation}\label{loss}
    \mathcal{L}(\mathbf{x,z};\theta)=-\sum_{t=1}^T \mathrm{Cross\-Entropy}(\hat{y}_t(x,z,y_{t-1}), y_t; \theta)
\end{equation}

where $\mathbf{x}$ denotes a text converted from human's speech, $\mathbf{z}$ denotes gesture context features. We use cross-entropy loss to measure the distance between each symbol in predicted and target LTL formulae respectively.

\begin{figure}[ht]
\centering
\includegraphics[width=1.\columnwidth]{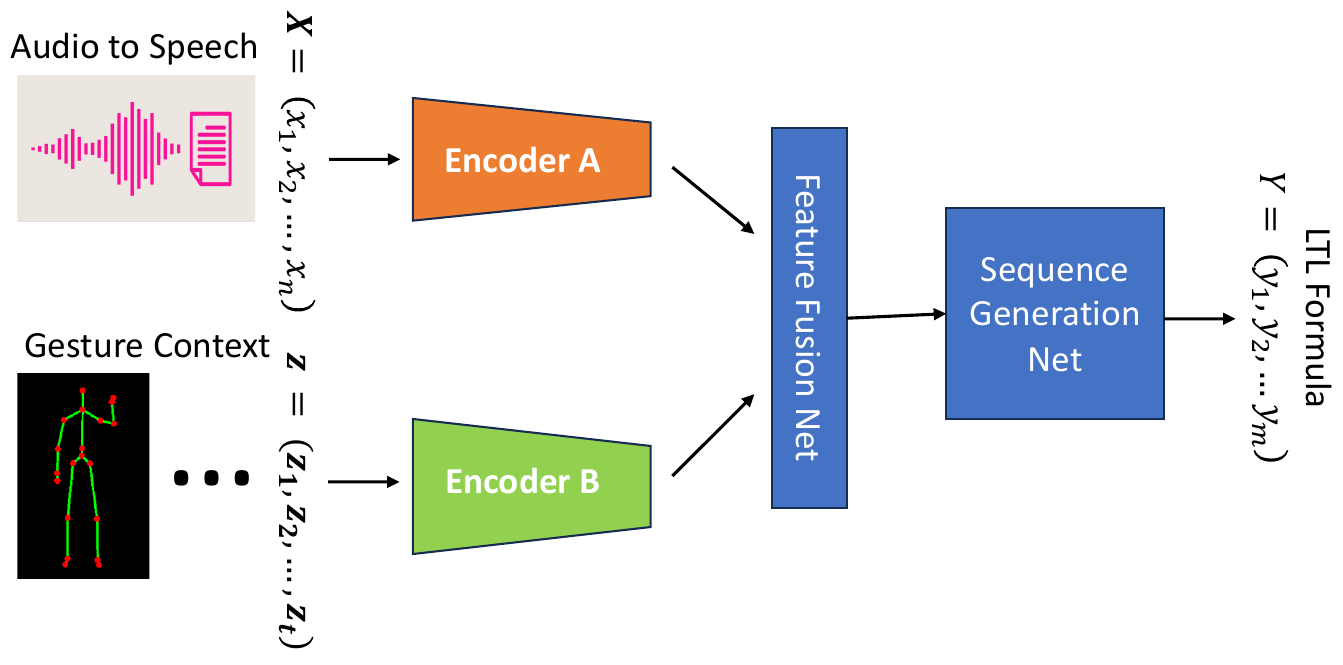}
\caption{The learning framework for translating a pair of speech and gesture data to an LTL formula that can solve multi-modal human task understanding problems.}
\label{fig:lang2ltl_arch}
\end{figure}


\subsection{Approach}
Our approach, depicted in Fig. \ref{fig:lang2ltl_arch}, entails a framework for acquiring the ability to forecast an LTL formula from a combination of speech and gesture. The architecture employs a two-stream network that is fed both a word sequence (speech) and a series of skeletal representations (gesture context). This dual input is then processed to yield a sequence of LTL symbols as output. Our training process involves optimizing the network through a cross-entropy loss, in accordance with Eq. \ref{loss}. The Encoder A is implemented by using a pretrained large language model BART \cite{lewis2019bart}. The Encoder B is pretrained by encoding and reconstructing our gesture data.


\section{Experiments} \label{sec:eval}
\label{sec:benchmark_tasks}

The \textit{NatSGD} dataset serves as a valuable resource for enhancing the learning of human-robot interaction tasks. In our study, we focus on evaluating the problem of multi-modal human task understanding. This evaluation serves to underscore the utility of the \textit{NatSGD} dataset and to emphasize the significance of harnessing multi-modal data in the process of training human-robot interaction systems.

\saketh{
\subsection{Baselines and Settings}

There exists no previous work that specifically addresses the task of learning a mapping from combined natural language sentences and gestures to an LTL formula. The closest work are \cite{wang2021learning,liu2022lang2ltl}, which tackles the challenge of translating natural language sentences into LTL formulas. Given this unique context, we establish two baseline models for comparison, each focusing on using either speech or gestures independently.

\subsubsection{Jaccard Similarity for Logic Formula Evaluation}
 Relevant works in logic formula inference have used Jaccard Similarity \cite{li2021explaining,saveri2023towards} as their evaluation metrics. Jaccard Similarity, defined as $J(A, B) = \frac{|A \cap B|}{|A \cup B|}$, measures the similarity between two sets by comparing their intersection $|A \cap B|$ to their union $|A \cup B|$. In evaluating logic formula prediction, Jaccard Similarity quantifies the overlap between the predicted and actual logic formulas, providing a robust assessment by considering the shared elements (true positives) and disregarding non-overlapping elements (false positives and false negatives).

\subsubsection*{Spot Scores for LTL Formula Equivalence}
In addition to Jaccard Similarity, we introduce a novel accuracy metric that tests the equivalence between two LTL formulas using the Spot library \cite{duret2016spot, duret2022spot}. The Spot library functions by converting input \(f\) and \(g\) (as well as their negations) into four automata: \(A_f\), \(A_{\neg f}\), \(A_g\), and \(A_{\neg g}\). It then ensures that both \(A_f \otimes A_{\neg g}\) and \(A_g \otimes A_{\neg f}\) are empty, validating formula equivalence. This metric, previously utilized in \cite{liu2022lang2ltl}, provides a precise measure for evaluating LTL formula predictions.

To determine if predicted and ground-truth formulas \(f_1\) and \(f_2\) are equivalent, we employ the \texttt{spot.are\_equivalent()} function, illustrated in the following link\footnote{\url{https://spot.lre.epita.fr/tut04.html}}. We provide a sample Python code snippet for reference:
}
\begin{verbatim}
import spot

# Define two Spot formulas
f1 = spot.formula("GF(a & Xb)")
f2 = spot.formula("G(Fa & Fb)")

# Check if the formulas are equivalent
if spot.are_equivalent(f1, f2):
    print("Equivalent.")
else:
    print("Not equivalent.")
\end{verbatim}

Here, \texttt{f1} and \texttt{f2} represent the Spot formulas under comparison. The \texttt{spot.are\_equivalent()} function returns \texttt{true} if the formulas are equivalent, and \texttt{false} otherwise. By assessing the equivalence of all pairs of predicted and annotated LTL formulas, we calculate the Spot Score:

\begin{equation}\label{eq:spot-score}
    \text{Spot-Score} = \frac{\sum_{i=1}^N \texttt{spot.are\_equivalent}(f_i, g_i)}{N}
\end{equation}

\subsection{Results and Analysis}

We conduct a comprehensive assessment, combining quantitative and qualitative analyses, to underscore the significance of incorporating speech and gestures in Multi-Modal Human Task Understanding tasks. In this endeavor, we introduce a comparative analysis between two state-of-the-art large language models (LLMs): T5, developed by Google, and BART, created by Meta. This comparison aims to evaluate their efficacy in predicting Linear Temporal Logic (LTL) formulas from inputs comprising either speech alone or a combination of speech and gestures.

Tables \ref{tab:table1} and \ref{tab:table2} showcase the performance metrics for these models, illustrating the distinct benefits of leveraging both speech and gestures for a deeper understanding of tasks articulated by humans. Our findings indicate that integrating speech with gestures not only enhances model performance but also significantly outperforms the results obtained by relying on a single modality.

\begin{table}[h] \label{table:BART_spot}
    \centering
    \begin{tabular}{|l|c|r|}
        \hline
        \textbf{Model (using BART \cite{lewis2019bart})} & \textbf{Jaq Sim$\uparrow$} & \textbf{Spot Score$\uparrow$} \\
        \hline
        Speech Only & 0.934 & 0.434 
        \\
        \hline
        Gestures Only & 0.922 & 0.299 
        \\
        \hline
        Speech + Gestures & \textbf{0.944} & \textbf{0.588} 
        \\
        \hline
    \end{tabular}
    \caption{Performance Comparison: Jaccard Similarity and Spot Score results for LTL formula prediction using BART}

    \label{tab:table1}
\end{table}

\begin{table}[h]
    \centering
    \begin{tabular}{|l|c|r|}
        \hline
        \textbf{Model (using T5 \cite{raffel2020exploring})} & \textbf{Jaq Sim$\uparrow$} & \textbf{Spot Score$\uparrow$} \\
        \hline
        Speech Only & 0.917 & 0.299 
        \\
        \hline
        Gestures Only & 0.948 & 0.244 
        \\
        \hline
        Speech + Gestures & \textbf{0.961}  & \textbf{0.507}
        \\
        \hline
    \end{tabular}
    \caption{Performance Comparison: Jaccard Similarity and Spot Score outcomes for LTL formula prediction using T5}

    \label{tab:table2}
\end{table}

Furthermore, the juxtaposition of T5 and BART models in our analysis yields additional insights into the nuanced ways different Language Model architectures process and interpret multi-modal data. The comparative results presented in Tables \ref{tab:table1} and \ref{tab:table2} affirm the premise that, although speech is the predominant mode of communication, the inclusion of gestures provides indispensable contextual support, enhancing the interaction model's comprehension capabilities. This observation aligns with the statistical insights detailed in Appendix \ref{sec:insights} of our dataset.

\section{Conclusion}

In this study, we introduce the \textit{NatSGD} dataset, a comprehensive multimodal Human-Robot Interaction (HRI) resource encompassing unstructured human commands conveyed through speech and natural gestures, accompanied by synchronized demonstration trajectories. We provide an in-depth description of our experimental methodology, meticulously crafted to capture genuine human behavior during interactions with a robot while performing complex household tasks. 

The \textit{NatSGD} dataset is enriched with detailed annotations, facilitating the training of a wide range of household robotic tasks. This renders it an invaluable asset for advancing research and development in Human-Robot Interaction (HRI) and robot learning. The dataset supports research in various domains, encompassing Multimodal Perception, Visual (object, activity, event, etc.) Recognition, Imitation Learning, Reinforcement Learning from Demonstrations, and Learning-based Plan Recognition.


\bibliography{output}
\bibliographystyle{IEEEtran.bst}

\clearpage

\appendix
\section{Appendix}

\subsection{Participants}
\label{appen:participants}
Eighteen volunteers participated in this experiment (9 male and 9 female) and received cash of \$10 each. Participants were recruited through posting flyers around the university and using local mailing lists. Participants’ ages ranged from 18 to 31 years (Mean 20.91\textpm3.75). Their personalities were generally characterized as follows: Extroversion (5.56\textpm2.09), Agreeableness (9.17\textpm1.15), Conscientiousness (8.17\textpm1.38), Emotional Stability (7.56\textpm1.79), and Openness to Experience (7.89\textpm1.47). Equal distribution of technical to non-technical background participants were chosen based on our questionnaire and post interview where we assessed their exposure to robots, remote controlled or gesture controlled games or toys, and their whether their education or profession was considered to have a technology focus such as computer science and engineering. None of the participants had ever interacted with any type of robots before they participated in the study.

\subsection{Lab Setup}
\label{appen:robot_setup}
\begin{figure}[ht]
\centering
\includegraphics[width=\columnwidth]{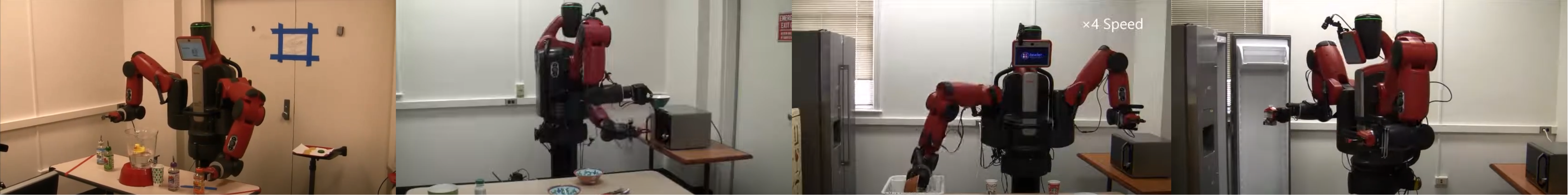}
\caption{Video Demonstration of the robot performing kitchen tasks (a) mixing drinks (b) microwaving bowl (c) cleaning up (d) fetch milk from fridge.}
\label{fig:lab_demo}
\end{figure}

\begin{figure}[ht]
\centering
\includegraphics[width=\columnwidth]{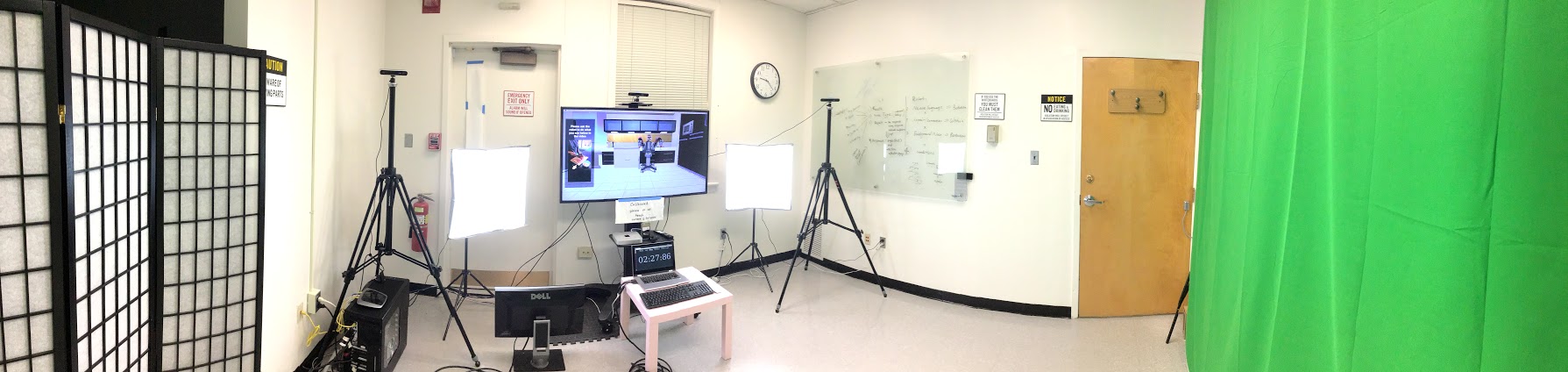}
\caption{Experiment Layout top view and Panorama image of the lab setup}
\label{fig:lab_setup}
\label{fig:experiment_setup}
\end{figure}

Participants were invited to the lab where we showed the Baxter humanoid and video recording of the robot performing kitchen actions that demonstrated the robot's ability to fetch items from a refrigerator, microwave a dish, and prepare a salad (see Fig. \ref{fig:lab_demo}). In addition, an we showed another video of an interaction between Baxter and a human where Baxter is following commands given by pointing or other demonstrations made by the body. We explained that since these videos, we have made a lot of improvements and wanted to understand how well the robot works with people and how well people can work with the robot. To be more believable, we explained that the robot motors were slow, so for the timing constraint, we created a virtual version of the robot that has the same AI engine. This allowed the participants to interact with virtual Baxter through a large monitor (55") at a 7-10' distance as shown in Fig. \ref{fig:experiment_setup}. Data was recorded from three Stereolabs ZED cameras from the front (camera 1), left and right at at approximately 30 degrees angle, at a resolution of 1280$\times$720 px. (720p) at 30 frames per second (FPS). A single iPhone 6S also recorded the participants audio and video in portrait mode at at a resolution of 720p at 30 FPS from approximately 15 degrees angle.

\subsection{Experiment Procedures}
\label{appen:experiment_procedures}

\label{appen:experiment_procedures:signup}
\textit{i) Participant Selection:} Participant filled out a screening questionnaire to sign up from university and local community advertisements. We collected information about their demographics, contact info for scheduling, language proficiency, hearing and vision deficiencies, and dominant hand information. The minimum criteria for selection was that the participants needed to be over 18 years of age (adult) native English speakers with normal hearing and vision (corrective lens or hearing aids were acceptable). Participants were also selected to have as diverse as possible from the sign up pool in gender, age, handedness, and ethnicity.

\label{appen:experiment_procedures:briefing}
\textit{ii) Briefing} After welcoming and a short introduction of the lab, the participants signed consent forms if they wish to continue. We showed recording of the past demonstrations of the humanoid performing various kitchen tasks. They are explained that due to the slow movement of the physical robot, they will be interacting with a simulated version of the robot with the same artificial intelligent brain of the robot they see in the lab.

\label{appen:experiment_procedures:simulator}
\textit{iii) Simulator} As shown in Fig. \ref{fig:natsgd_multi_cam_views}, the \textit{NatSGD} Robot Simulator was built using Unity 3D \cite{Unity_Technologies_undated-qy} with modified ROS plugin \cite{Bischoff_rossharp}. The system runs on a Intel i7 Gen 16GB RAM connected to the 55" TV. On the top right corner of the screen, a camera feed of the participant is overlayed as shown in Fig \ref{fig:intro}. This serves as a feedback mechanism to the participants to help them stay within the frame. The simulation runs based on ROS storyline module that we developed that helps us design the workflow of the recipe in the Unity environment. ROS messaging can (a) control camera angles for staging the desired point of view of the scene and the robot (b) the video instructions of the step of the recipe (c) robot responses such as head nodding, confusion face, and normal blinking face, and (d) robot actions such as navigation or performing cooking tasks. The robot navigation path was predetermined as there were limited number of target locations while the real-time inverse kinematics (IK) of the head nods and robot arms were implemented with BioIK \cite{Starke2017-cx, Starke2018-pb}. To wait for participant command, the robot looks at the participant. Before performing each task, the robot looks at the target object(s) to demonstrate robot's attention. 


\label{appen:experiment_procedures:practice_session}
\textit{iv) Practice Session} During the practice session, participants interact with the robot to navigate the robot in discrete steps in the kitchen and perform steps to cut an apple. The practice session lasts roughly 5 minutes until the participant feels comfortable. During practice phase, the participants can interact with the research facilitator. After the practice phase, once the data collection begins, they are not allowed to ask any questions until the data collection session was over.

\textit{v) WoZ Control Policy} For both practice and the experiment sessions, the researcher facilitator (wizard) controls the workflow to move ahead as long as the participant's command is related to the task at hand and is discernible. Robot nods ``yes" (head up and down motion) on commands that are understood. For unintelligible or unrelated commands, the robot displays the confusion face. For example, if a participant gestures to move right, or say move right, the wizard makes the robot move right. However, if the participant mumbles and the wizard cannot hear the participant, the wizard prompts with a confused robot face for clarification from the participant. The clarification questions are designed to be ambiguous so the participant attempts different strategies in providing instructions for the same task. The participant is allowed a maximum of five attempts per command. If unsuccessful after that, the task is skipped. In rare cases, the task can also be skipped, if the Unity game becomes frozen or Baxter is not able to solve the IK for grasping objects for more than 30 seconds.

\label{appen:post_survey_and_interview}
\textit{vi) Post Survey and Interview} After the data collection, participants are asked to fill out a survey providing their impression of the robot, their own personality, and the workload experienced. We finally wrap up the session with a semi-structured interview gathering additional details about their experience and debriefing them of any information we shared that was used to deceive them.


\subsection{Background Noise}
\label{appen:background_noise}
One hypothesis was that background noise can cause people to use more gestures. We considered 3 types of noise recording playback (lawn mower, people talking, and music), but only tested with people talking as background noise as that was the only example people found to be believable and not simulated. We tested at 3 sets of loudness (M (dB) = 58, 63, 70, SD = 10, 13, 15). In our study (N=8), from people's use of speech and gesture and the post-interview, we found that (a) people tune out the background noise, instead of using more gestures or, (b) people wait for gaps of silence or lower level noise in cases of speech or periodic noise, and (c) the noise had to be so loud that none of the speech can be heard at all for them to use gestures instead of speech. For these reasons, we decided not to use background noise as an independent variable.

\subsection{Robot Face and Name}
\label{appen:robot_face_and_name}
To reduce the affect of perceived gender, age, and personality by manipulating facial attributes, we considered the 17 face dimensions based on \cite{Kalegina2018-ki} study to  design the face of the robot to be the most neutral face. The mouth of the robot was removed as not having a mouth did not have significant adverse effect on the neutral perception of the robot.  Having a mouth seemed to give people the idea  that the robot could speak, potentially causes the participant to prefer speech over gesture. For the robot to appear dynamic, friendly, and intelligent, we made the robot blink randomly between 12 and 18 blinks per minute \cite{Takashima2008-tz} with ease-in and ease-out motion profile \cite{Trutoiu2011-se, thomas1995illusion}. We further conducted pilot tests to analyze the head nod motions (velocity and number of nods) and facial expressions for confusion expression. Additionally, we avoided using gender specific pronouns ``he/him" and ``she/her" and referred to the robot as ``the robot" or ``Baxter" which is also the manufacturer given name printed on robot body that tends to be used both as a male and female name \cite{BaxterName_Wikipedia-qz}.

\subsection{Practice Session}
\label{appen:practice_session}
During practice, it is important to make sure that participants are not primed to use one modality versus the other. So steps were taken to design the session with a mixture of related and unrelated commands where both speech and gestures were used to command the robots. If participants used a single modality only, they were encouraged to test out using the other modality. Participants interacted with the robot and asked researchers questions during practice. Once the practice was completed, participants were not allowed to interact with anyone other than the robot even if they had questions or felt stuck as they were told that the experiment was designed for them to experience such scenarios and had to use creative methods make the robot understand what they wanted the robot to do.

\subsection{WoZ Clues}
\label{appen:woz_cues}
People can be quite intuitive in figuring out the patterns such as key press and mouse click sounds corresponding to robot actions. We experimented with masking the actual clicks and key presses with random ones. However, in the post interview the pilot test participants still seem to be able to figure out that researchers might be controlling the robot. So we created a soft rubber remote control keys that use IR receiver using Arduino micro-controller USB adapter to send keys to the WoZ UI with virtually no sound that the researcher keeps in their pocket. With this implementation, during the experiment, the researchers made sure when the experiment is being conducted, they do not sit at the control computer and appear to be moving around doing other things appearing busy, staring at their phone seemingly distracted, or looking at the participants showing attention in making sure the system was working without any technical issues. With this implementation, 100\% of the participants believed that the robot was acting on its own and none of the participants suspected the WoZ setup to be a possibility.

\subsection{Instructions}
\label{appen:instructions_human}
Based on the recommendations \cite{Fothergill2012-sb}, we tested various modalities for our applications. Our findings in our pilot studies were in-line with \cite{Fothergill2012-sb, Charbonneau2011-yi} where the instruction modality had a significant impact on the participants' behavior. For instances where text instructions were provided similar to \cite{Cauchard2015-vv}, participants preferred speech and used the exact words for the action and the object with little or no gestures. With videos of people performing the task similar to \cite{Charbonneau2011-yi}, participants copied the exact style of the demonstration of the actor. The one with the most variance in speech vocabulary and styles of gestures were when we showed before-after video clips to show the pre-task and post-task states, for example, to turn on a stove, we showed a zoomed in video of a stove that was turned off, and faded out to a video of the stove with fire burning. For cutting apple, video of a whole apple on a cutting board being approached by a knife and faded into apple that was cut into pieces where the knife is leaves the screen. And these videos were repeated in a loop with a 1 second gap in between.

\subsection{An Example Scenario}
\label{appen:storyline}
\saketh{
The setup for this experiment was designed to allow participants to engage naturally with a robot by providing it with detailed, step-by-step instructions for preparing food in a kitchen setting. The sequence of food preparation tasks was intentionally developed to include repetitive actions across different objects to enhance the interaction's complexity and realism. This approach introduced a dynamic component to the experiment, where the robot alternated between executing commands immediately and asking participants for further clarification or additional instructions, adding an element of unpredictability to the task execution.

The experiment revolved around the preparation of a vegetable soup recipe, which included a variety of culinary tasks spanning from the initial preparation to the final serving steps. For prepping, the ingredients had to be fetched and cut. The cooking steps included turning a gas stove burner on and off, sauteing, stirring, seasoning, transferring, and covering/uncovering a pot. Serving the soup required fetching a bowl and pouring the soup into the bowl and placing it on the counter. The final steps included cleaning up by putting away the pots and pans into the sink.
}

\subsection{Post Processing: Data Processing, Synchronization, and Camera Calibration}
\label{appen:post_processing_sync_cam_calib}
It is important to clean up the data and make it easy for researchers to use the data. Careful iterative steps were taken to prepare the data to ensure integrity, quality, validity, and fairness. The raw data is processed, annotated, validated, visualized, and curated for downstream analysis and machine learning tasks in the following way.

\textit{i) Multi-camera Calibration}: A standard 12$\times$8 5" checker board was recorded using ROS, and Kalibr package \cite{furgale2013unified} to compute the cameras intrinsic and extrinsic matrix. If the average re-projection error was greater than 1 px., the calibration was repeated.

\textit{ii) Multi-camera Audio-Video Synchronization and Data Compression}: All the data was recorded using ROS bag. These recorded video frames from each cameras tend to have dynamically varying frames per second rates anywhere from 25 fps to 32 fps which makes it difficult to synchronize with sound. For this reason, the audio recording is extracted from a well established audio-video camera such as Apple iPhone camera. A flashing color screen from another computer is placed in the middle of the lab within all the cameras' field of view. A ROS start message is also published to store and identify the starting flag of the session for all other data.
The changing color from red to blue is used to denote the mark of the starting frame and the ROS bag start message time is used for offsetting other messages. The frames are then streamed to a canvas that is 6$\times$ the size of 720p i.e., 2560$\times$2560 where each row is a 720p stereo camera frame. At 33.33ms the latest state of the frame is recorded. The iPhone video is also clipped starting from the blue frames whose sound is then merged with the large canvas video to generate the data. This data is then re-encoded to be compressed using FFMPEG and NVIDIA TITAN X H.264 encoder \cite{ffmpeg-hwaccelIntro-th}. 

\subsection{Data Annotation}
\label{appen:data_annotation}
Similar to \cite{Liu2021-ph}, multiple people annotated the data for purposes of speed and reliability. Each annotations were annotated at least twice with up to three round of agreement checks. Data with difference in opinion that could not come to agreement were subject to voting by five annotators to require minimum 80\% score, otherwise were left out from the final label list. The data was annotated using Fast Event Video Annotation Tool (FEVA \cite{shrestha2023feva}) (see Fig. \ref{fig:feva_p40}). Speech and gesture onset and offset of each command is annotated by two researchers. The data is stored using a \texttt{json} format with FEVA dataset schema v2.1. Speech was generated using Otter \cite{Otter-asr-vw} which was audited by two researchers. The human pose were extracted using OpenPose \cite{Cao2019-iv} which we filtered to discard frames with large errors or joints that were not detected. Each event is then annotated by five independent annotators using FEVA crowd based on the ontology as shown in Fig. \ref{fig:natcom_dataset_tree}. 

\subsection{Inter-rater Reliability}
\label{appen:interrater_reliability}
Inter-rate reliability (IRR) was computed using Cohen's Kappa \cite{cohen1960coefficient} and annotation was updated in three rounds where 100\% agreement was reached. The round 1 IIR for modality was 99.7\% for speech, 94.4\% for gesture. Similarly, for speech reference for object was 97.3\%, for action was 99.3\%. For gesture type, Task oriented was 85.2\%, and non-intent was 72.0\%. For gesture reference for object was 77.9\%, and for action 80.3\%. For task object reference body parts was 91.4\%, task action body parts 88.8\%, and non-intent body parts was 79.9\%. All scoring well above recommended 70\%. Modality and speech reference reached 100\% agreement in round 2 correcting for mistakes and revised look. For gesture type, reference gesture, and gesture body parts, most reached approximately 90\% in round 2 and 100\% in round 3.

\subsection{Dataset Statistical Insights} \label{sec:insights}

\textit{Insight 1} Overall 97.3\% of the commands contained speech and 81.5\% contained gesture, with 81.0\% containing both at the same time. However, commands with speech but without gesture drops to 18.46\% and gesture only to 2.7\%. This implies much higher preference for speech over gesture, but also shows gesture being used with speech significantly.

\textit{Insight 2}: While gestures are used 81.54\% of the time, it is only used 2.7\% of time implying people are less likely to use gesture independently in the context of kitchen tasks. When gestures are used, 74.68\% of the time was conveying task oriented messages while 56.55\% of the time also contained non-intent gestures. Independently, task oriented message that did not have any non-intent gestures were 42.70\% of the time. This implies non-intent gestures can be confounding the overall gesture and wrongly interpreted by models that naively model the human motion. For instance, there are several instances where some participants use a rhythmic (beat) gestures to regulate and help them express a message they are having difficulty expressing. Those instances, naively the gesture look a cutting motion. Without speech these gestures are ambiguous and can be misleading.

\begin{figure}[ht]
\centering
\includegraphics[width=\columnwidth]{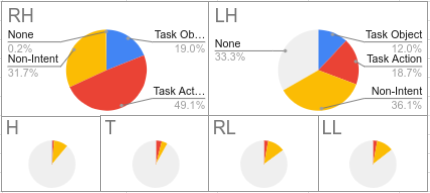}
\caption{The distribution of gesture by body part usage for intent and reference dimensions for the gesture command. Right hand (RH), Left hand (LH), Head (H), Torso (T), Right leg (RL), and Left leg (LL).}
\label{fig:dataset_gesture_action_object_active_passive_bodyparts}
\end{figure}

As shown in Fig. \ref{fig:dataset_gesture_action_object_active_passive_bodyparts} majority of the time, right and left hands are being for gestures which seems obvious. However, regardless of the handedness, in a two-tailed T-test across participants, people on average used left hand more to indicate action gestures (M=63.50, SD=18.87) compared to (M=25.44, SD=13.78) with the right hand t=6.90933, \textit{p\textless0.00001}. This can be explained by the fact that, most of the actions with the left hands were for activities that was happening to the left side of the user implying spatial priority people give to express with gesture.

\textit{Insight 3}: The data format that we collected can be used to train machine learning models to use multimodal data for many tasks such as command recognition, visual question and answering, temporal segmentation of people behavior, learning cascaded command intents, and so on. These can be useful for teaching robots to understand our intents of our commands better.

\textit{Insight 4}: As stated in \textit{Insight 2} and in the speech data, people often refer to spatial objects and locations, items referred to from previous commands, using the same words for two items such as pot for both the pan and the pot, using words like fire to mean the stove and so on. These kinds of information is required by the robot to make observation outside of the human command to understand the bigger context of the commands. This dataset take a step in that direction. But more work is needed.

\begin{itemize}
    \item \textit{Modality}: The first distinction is the modality of the communications with the presence or absence of speech and gesture. Speech includes primarily the language, while para-linguistic features are reserved for future work. Gestures include full body static postures or dynamic movements, however, facial expressions and eye gaze are planned for future work.
    
    \item \textit{Reference}: For robots to perform tasks, information of the \textit{action} and the target \textit{object} is key for the competition of the said task \cite{yang2015robot}. Therefore, each \textit{speech} and \textit{gesture} clips are independently annotated for their references to the \textit{objects} that the task requires the robot to interact with and the \textit{actions} the robot needs to perform with or to the \textit{objects}. These provide rich information on the ``what", ``where", and the ``how" of the task.
    
    \item \textit{Gesture Type}: Building on the Communicative-Informative dichotomy \cite{Abner2015-ot} introduced by Lyons in Semantics \cite{lyons1977semantics}, we segment the gestures based on the \textit{intent} of the speaker as perceived by the receiver. We segment gestures into \textit{Task} oriented and \textit{Non-intent}. \textit{Task} gestures serves to clearly and intentionally communicate the desired task to be performed. \textit{Non-intent} gestures include static postures or dynamic movements that are not intended to communicate the specifics of the desired task. These gestures may provide additional contextual information, however, they have to be implied by the receiver and often tend to have a large degree of disagreement between annotators on what they mean or imply. Isolating these motions can help machine learning systems to classify gesture useful for interpreting the desired task information the speaker is intending to communicate from the gesture along with the speech. The \textit{non-intent gestures} can also be used to gather the state of the speaker.

    \item \textit{Body Parts}: For all task action or object or non-intent gestures, the corresponding body parts are clearly annotated and are divided into 6 body part groups: (1) \textit{head} (2) \textit{torso} (3) \textit{left hand} (4) \textit{right hand} (5) \textit{left leg} and (6) \textit{right leg}.
\end{itemize}

\subsection{Ground Truth Labels}
\label{sec:ground_truth_labels}
The structure of this dataset has been from a utilitarian perspective for robotics and machine learning applications. \textit{NatSGD} contains labels for each task that were completed, for eg. cutting an onion or a tomato. For each task, we also break them down into labels of the their sub-task such as grabbing the knife, holding the onion, and cutting are provided. We include synchronized data from both robot and human perspectives.

\begin{figure}[ht]
\centering
\includegraphics[width=\columnwidth]{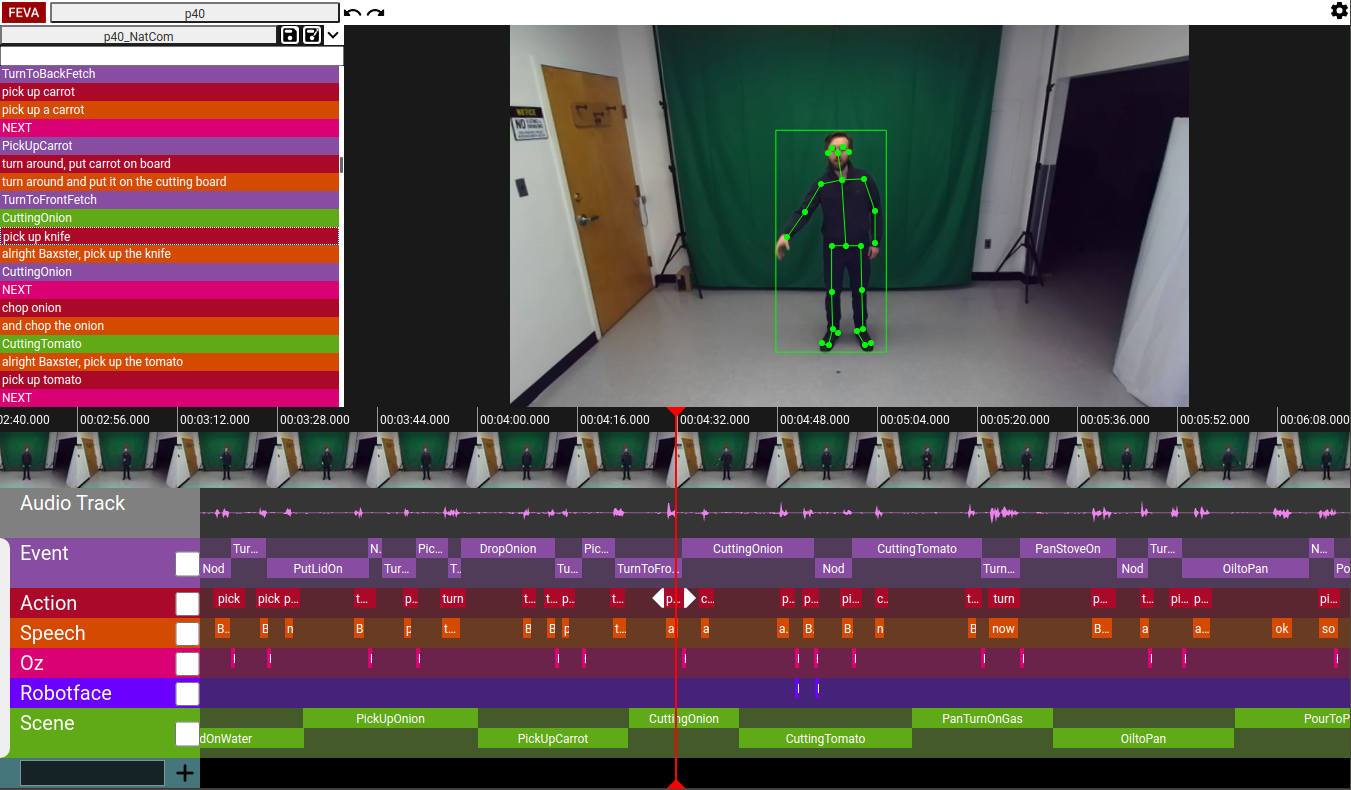}
\caption{Example of temporal annotation of the tasks, sub-tasks, speech, and gesture of participant p40 with Fast Event Video Annotation (FEVA \cite{shrestha2023feva}) tool.}
\label{fig:feva_p40}
\end{figure}

On the human side, we temporally segmented human commands as shown in Fig. \ref{fig:feva_p40}. Based on \textit{modality}, each command was annotated for consisting relevant \textit{speech} and \textit{gesture}. We then further distinguished them based if they \textit{referred} to \textit{objects} or \textit{action}. The gesture is annotated as containing task-related (\textit{intentional}) and task-unrelated (\textit{unintentional}) gestures. Finally, for all gestures, we also annotated the role of each \textit{body part} in task specific action/ object reference and non-intent movements as illustrated in Fig. \ref{fig:natcom_dataset_tree}. Please see appendix for more details in Appendix. \ref{sec:ground_truth_labels}. 

\begin{figure}[ht]
\centering
\includegraphics[width=0.9\columnwidth]{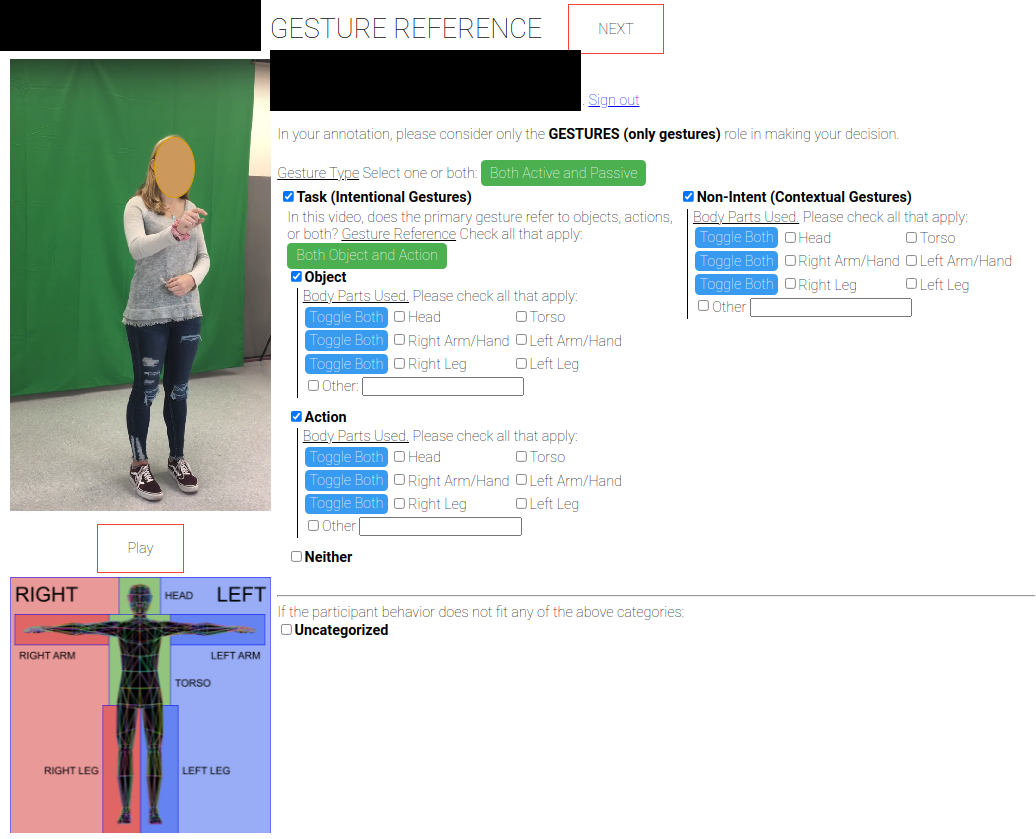}
\caption{Gesture Type, Reference, and Body Parts Annotation of participant p61 using FEVA Crowd tool.}
\label{fig:feva_p61}
\end{figure}

\begin{figure}[ht]
\centering
\includegraphics[width=0.7\columnwidth]{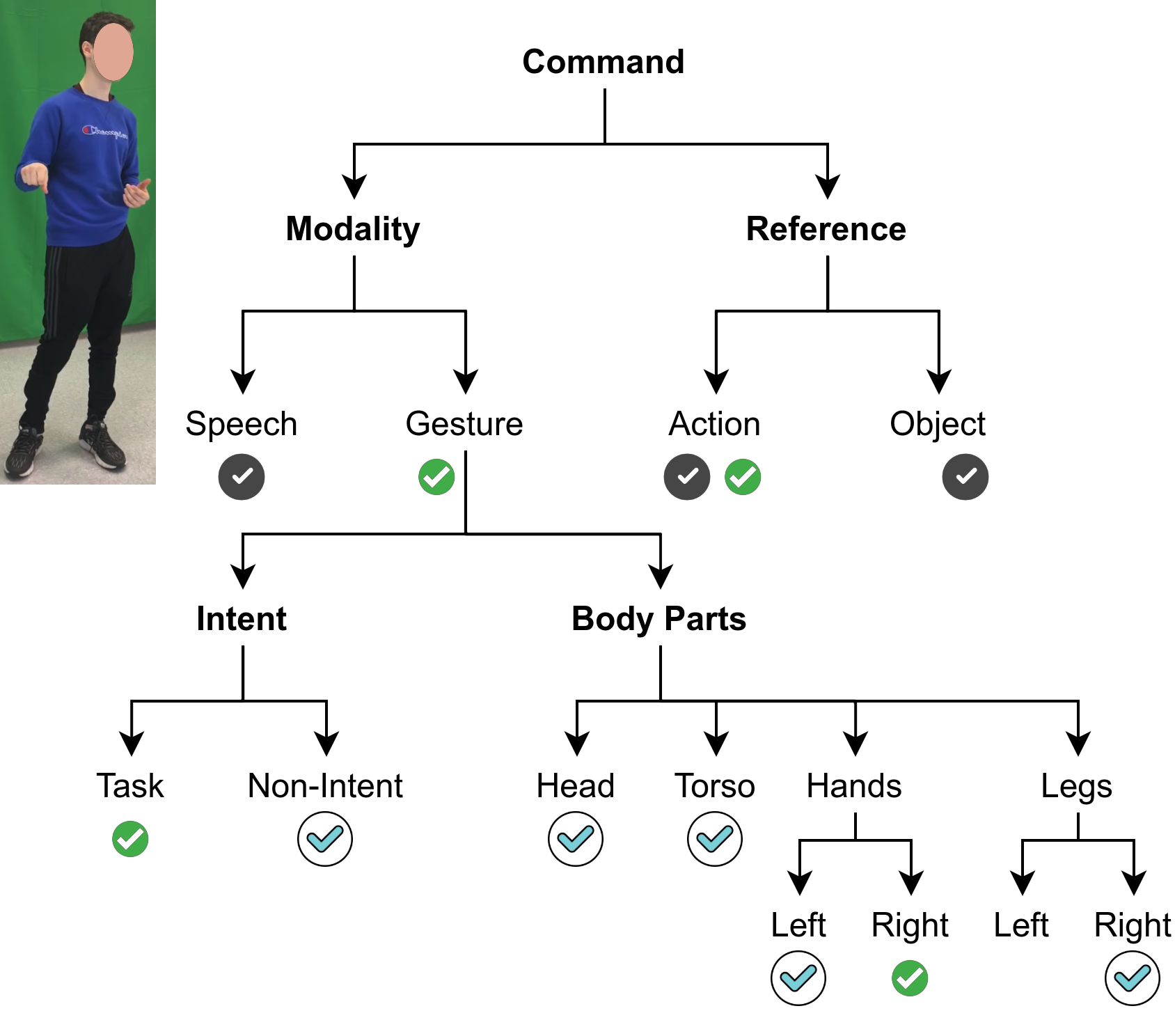}
\caption{In the example above, participant p58 says ``Baxter, stir the pot", with right hand showing the stirring gesture while the body shifts to his left with head and left hand moving that could be read as being unsure. Here 'stir' is speech action, 'pot' is speech object, while right hand is gesturing for stirring task action, the head, torso, left hand and right leg are unintentional gestures.}
\label{fig:natcom_dataset_tree}
\end{figure}

On the robot side, sequences of images are temporally synchronized across all cameras along with the audio. The depth images are generated by using the normalized distance from the robot's egocentric view. Additionally, individual object's unique segmentation and their object semantic segmentation, based on their object categories such as food, utensils, and appliances are also provided. \textit{NatSGD} also includes expert trajectories of the end-effectors, head, and the robot base that demonstrates how to perform the tasks, head control, and robot navigation.

\bibliography{output}

\begin{thebibliography}{10}
\providecommand{\url}[1]{#1}
\csname url@samestyle\endcsname
\providecommand{\newblock}{\relax}
\providecommand{\bibinfo}[2]{#2}
\providecommand{\BIBentrySTDinterwordspacing}{\spaceskip=0pt\relax}
\providecommand{\BIBentryALTinterwordstretchfactor}{4}
\providecommand{\BIBentryALTinterwordspacing}{\spaceskip=\fontdimen2\font plus
\BIBentryALTinterwordstretchfactor\fontdimen3\font minus
  \fontdimen4\font\relax}
\providecommand{\BIBforeignlanguage}[2]{{%
\expandafter\ifx\csname l@#1\endcsname\relax
\typeout{** WARNING: IEEEtran.bst: No hyphenation pattern has been}%
\typeout{** loaded for the language `#1'. Using the pattern for}%
\typeout{** the default language instead.}%
\else
\language=\csname l@#1\endcsname
\fi
#2}}
\providecommand{\BIBdecl}{\relax}
\BIBdecl

\bibitem{tomasello2010origins}
M.~Tomasello, \emph{Origins of human communication}.\hskip 1em plus 0.5em minus
  0.4em\relax MIT press, 2010.

\bibitem{novoa2017multichannel}
J.~Novoa, J.~P. Escudero, J.~Fredes, J.~Wuth, R.~Mahu, and N.~B. Yoma,
  ``Multichannel robot speech recognition database: Mchrsr,'' \emph{arXiv
  preprint arXiv:1801.00061}, 2017.

\bibitem{james2018open}
J.~James, L.~Tian, and C.~I. Watson, ``An open source emotional speech corpus
  for human robot interaction applications.'' in \emph{Interspeech}, 2018, pp.
  2768--2772.

\bibitem{vasudevan2018object}
A.~B. Vasudevan, D.~Dai, and L.~Van~Gool, ``Object referring in visual scene
  with spoken language,'' in \emph{2018 IEEE winter conference on applications
  of computer vision (WACV)}.\hskip 1em plus 0.5em minus 0.4em\relax IEEE,
  2018, pp. 1861--1870.

\bibitem{narayan2019collaborative}
A.~Narayan-Chen, P.~Jayannavar, and J.~Hockenmaier, ``Collaborative dialogue in
  minecraft,'' in \emph{Proceedings of the 57th Annual Meeting of the
  Association for Computational Linguistics}, 2019, pp. 5405--5415.

\bibitem{padmakumar2022teach}
A.~Padmakumar, J.~Thomason, A.~Shrivastava, P.~Lange, A.~Narayan-Chen,
  S.~Gella, R.~Piramuthu, G.~Tur, and D.~Hakkani-Tur, ``Teach: Task-driven
  embodied agents that chat,'' in \emph{Proceedings of the AAAI Conference on
  Artificial Intelligence}, vol.~36, no.~2, 2022, pp. 2017--2025.

\bibitem{pisharady2015recent}
P.~K. Pisharady and M.~Saerbeck, ``Recent methods and databases in vision-based
  hand gesture recognition: A review,'' \emph{Computer Vision and Image
  Understanding}, vol. 141, pp. 152--165, 2015.

\bibitem{shukla2016multi}
D.~Shukla, {\"O}.~Erkent, and J.~Piater, ``A multi-view hand gesture rgb-d
  dataset for human-robot interaction scenarios,'' in \emph{2016 25th IEEE
  international symposium on robot and human interactive communication
  (RO-MAN)}.\hskip 1em plus 0.5em minus 0.4em\relax IEEE, 2016, pp. 1084--1091.

\bibitem{mazhar2018towards}
O.~Mazhar, S.~Ramdani, B.~Navarro, R.~Passama, and A.~Cherubini, ``Towards
  real-time physical human-robot interaction using skeleton information and
  hand gestures,'' in \emph{2018 IEEE/RSJ International Conference on
  Intelligent Robots and Systems (IROS)}.\hskip 1em plus 0.5em minus
  0.4em\relax IEEE, 2018, pp. 1--6.

\bibitem{chen2018wristcam}
F.~Chen, H.~Lv, Z.~Pang, J.~Zhang, Y.~Hou, Y.~Gu, H.~Yang, and G.~Yang,
  ``Wristcam: A wearable sensor for hand trajectory gesture recognition and
  intelligent human--robot interaction,'' \emph{IEEE Sensors Journal}, vol.~19,
  no.~19, pp. 8441--8451, 2018.

\bibitem{chang2019improved}
J.-Y. Chang, A.~Tejero-de Pablos, and T.~Harada, ``Improved optical flow for
  gesture-based human-robot interaction,'' in \emph{2019 International
  Conference on Robotics and Automation (ICRA)}.\hskip 1em plus 0.5em minus
  0.4em\relax IEEE, 2019, pp. 7983--7989.

\bibitem{gomez2019caddy}
A.~Gomez~Chavez, A.~Ranieri, D.~Chiarella, E.~Zereik, A.~Babi{\'c}, and
  A.~Birk, ``Caddy underwater stereo-vision dataset for human--robot
  interaction (hri) in the context of diver activities,'' \emph{Journal of
  Marine Science and Engineering}, vol.~7, no.~1, p.~16, 2019.

\bibitem{neto2019gesture}
P.~Neto, M.~Sim{\~a}o, N.~Mendes, and M.~Safeea, ``Gesture-based human-robot
  interaction for human assistance in manufacturing,'' \emph{The International
  Journal of Advanced Manufacturing Technology}, vol. 101, no.~1, pp. 119--135,
  2019.

\bibitem{nuzzi2021hands}
C.~Nuzzi, S.~Pasinetti, R.~Pagani, G.~Coffetti, and G.~Sansoni, ``Hands: an
  rgb-d dataset of static hand-gestures for human-robot interaction,''
  \emph{Data in Brief}, vol.~35, p. 106791, 2021.

\bibitem{de2021introducing}
J.~de~Wit, E.~Krahmer, and P.~Vogt, ``Introducing the nemo-lowlands iconic
  gesture dataset, collected through a gameful human--robot interaction,''
  \emph{Behavior Research Methods}, vol.~53, no.~3, pp. 1353--1370, 2021.

\bibitem{luan2016reliable}
W.~Luan, Y.~Yang, C.~Ferm{\"u}ller, and J.~S. Baras, ``Reliable attribute-based
  object recognition using high predictive value classifiers,'' in
  \emph{European Conference on Computer Vision}.\hskip 1em plus 0.5em minus
  0.4em\relax Springer, 2016, pp. 801--815.

\bibitem{matuszek2014learning}
C.~Matuszek, L.~Bo, L.~Zettlemoyer, and D.~Fox, ``Learning from unscripted
  deictic gesture and language for human-robot interactions,'' in
  \emph{Proceedings of the AAAI Conference on Artificial Intelligence},
  vol.~28, no.~1, 2014.

\bibitem{rodomagoulakis2016multimedia}
I.~Rodomagoulakis, N.~Kardaris, V.~Pitsikalis, A.~Arvanitakis, and P.~Maragos,
  ``A multimedia gesture dataset for human robot communication: Acquisition,
  tools and recognition results,'' in \emph{2016 IEEE International Conference
  on Image Processing (ICIP)}.\hskip 1em plus 0.5em minus 0.4em\relax IEEE,
  2016, pp. 3066--3070.

\bibitem{azagra2017multimodal}
P.~Azagra, F.~Golemo, Y.~Mollard, M.~Lopes, J.~Civera, and A.~C. Murillo, ``A
  multimodal dataset for object model learning from natural human-robot
  interaction,'' in \emph{2017 IEEE/RSJ International Conference on Intelligent
  Robots and Systems (IROS)}.\hskip 1em plus 0.5em minus 0.4em\relax IEEE,
  2017, pp. 6134--6141.

\bibitem{chen2022real}
H.~Chen, M.~C. Leu, and Z.~Yin, ``Real-time multi-modal human-robot
  collaboration using gestures and speech,'' \emph{Journal of Manufacturing
  Science and Engineering}, pp. 1--22, 2022.

\bibitem{Dahlback1993-el}
N.~Dahlb{\"a}ck, A.~J{\"o}nsson, and L.~Ahrenberg, ``Wizard of oz studies,'' in
  \emph{Proceedings of the 1st international conference on Intelligent user
  interfaces - {IUI} '93}.\hskip 1em plus 0.5em minus 0.4em\relax New York, New
  York, USA: ACM Press, 1993.

\bibitem{hofer2021sim2real}
S.~H{\"o}fer, K.~Bekris, A.~Handa, J.~C. Gamboa, M.~Mozifian, F.~Golemo,
  C.~Atkeson, D.~Fox, K.~Goldberg, J.~Leonard \emph{et~al.}, ``Sim2real in
  robotics and automation: Applications and challenges,'' \emph{IEEE
  transactions on automation science and engineering}, vol.~18, no.~2, pp.
  398--400, 2021.

\bibitem{lee2019talking}
G.~Lee, Z.~Deng, S.~Ma, T.~Shiratori, S.~S. Srinivasa, and Y.~Sheikh, ``Talking
  with hands 16.2 m: A large-scale dataset of synchronized body-finger motion
  and audio for conversational motion analysis and synthesis,'' in
  \emph{Proceedings of the IEEE/CVF International Conference on Computer
  Vision}, 2019, pp. 763--772.

\bibitem{kucherenko2021large}
T.~Kucherenko, P.~Jonell, Y.~Yoon, P.~Wolfert, and G.~E. Henter, ``A large,
  crowdsourced evaluation of gesture generation systems on common data: The
  genea challenge 2020,'' in \emph{26th international conference on intelligent
  user interfaces}, 2021, pp. 11--21.

\bibitem{krishnaswamy2020formal}
N.~Krishnaswamy and J.~Pustejovsky, ``A formal analysis of multimodal referring
  strategies under common ground,'' in \emph{Proceedings of the 12th Language
  Resources and Evaluation Conference}, 2020, pp. 5919--5927.

\bibitem{ahuja2020no}
C.~Ahuja, D.~W. Lee, R.~Ishii, and L.-P. Morency, ``No gestures left behind:
  Learning relationships between spoken language and freeform gestures,'' in
  \emph{Findings of the association for computational linguistics: EMNLP 2020},
  2020, pp. 1884--1895.

\bibitem{habibie2021learning}
I.~Habibie, W.~Xu, D.~Mehta, L.~Liu, H.-P. Seidel, G.~Pons-Moll, M.~Elgharib,
  and C.~Theobalt, ``Learning speech-driven 3d conversational gestures from
  video,'' in \emph{Proceedings of the 21st ACM International Conference on
  Intelligent Virtual Agents}, 2021, pp. 101--108.

\bibitem{kucherenko2021multimodal}
T.~Kucherenko, R.~Nagy, M.~Neff, H.~Kjellstr{\"o}m, and G.~E. Henter,
  ``Multimodal analysis of the predictability of hand-gesture properties,''
  \emph{arXiv preprint arXiv:2108.05762}, 2021.

\bibitem{habibie2022motion}
I.~Habibie, M.~Elgharib, K.~Sarkar, A.~Abdullah, S.~Nyatsanga, M.~Neff, and
  C.~Theobalt, ``A motion matching-based framework for controllable gesture
  synthesis from speech,'' in \emph{Special Interest Group on Computer Graphics
  and Interactive Techniques Conference Proceedings}, 2022, pp. 1--9.

\bibitem{hart2006nasa}
S.~G. Hart, ``Nasa-task load index (nasa-tlx); 20 years later,'' in
  \emph{Proceedings of the human factors and ergonomics society annual
  meeting}, vol.~50, no.~9.\hskip 1em plus 0.5em minus 0.4em\relax Sage
  publications Sage CA: Los Angeles, CA, 2006, pp. 904--908.

\bibitem{pennington2014glove}
J.~Pennington, R.~Socher, and C.~D. Manning, ``Glove: Global vectors for word
  representation,'' in \emph{Proceedings of the 2014 conference on empirical
  methods in natural language processing (EMNLP)}, 2014, pp. 1532--1543.

\bibitem{lewis2019bart}
M.~Lewis, Y.~Liu, N.~Goyal, M.~Ghazvininejad, A.~Mohamed, O.~Levy, V.~Stoyanov,
  and L.~Zettlemoyer, ``Bart: Denoising sequence-to-sequence pre-training for
  natural language generation, translation, and comprehension,'' \emph{arXiv
  preprint arXiv:1910.13461}, 2019.

\bibitem{abeyruwan2022sim2real}
S.~Abeyruwan, L.~Graesser, D.~B. D'Ambrosio, A.~Singh, A.~Shankar, A.~Bewley,
  and P.~R. Sanketi, ``i-sim2real: Reinforcement learning of robotic policies
  in tight human-robot interaction loops,'' \emph{arXiv preprint
  arXiv:2207.06572}, 2022.

\bibitem{kaspar2020sim2real}
M.~Kaspar, J.~D.~M. Osorio, and J.~Bock, ``Sim2real transfer for reinforcement
  learning without dynamics randomization,'' in \emph{2020 IEEE/RSJ
  International Conference on Intelligent Robots and Systems (IROS)}.\hskip 1em
  plus 0.5em minus 0.4em\relax IEEE, 2020, pp. 4383--4388.

\bibitem{Unity_Technologies_undated-qy}
{Unity Technologies}, ``Unity {3D} game engine,'' \url{https://unity.com/},
  2021, accessed: 2021-9-28.

\bibitem{Bischoff_rossharp}
M.~Bischoff, ``ros-sharp: {ROS\#} is a set of open source software libraries
  and tools in c\# for communicating with {ROS} from .{NET} applications, in
  particular {Unity3D},'' 2021.

\bibitem{Starke2017-cx}
S.~Starke, N.~Hendrich, D.~Krupke, and {others}, ``Evolutionary multi-objective
  inverse kinematics on highly articulated and humanoid robots,'' \emph{2017
  IEEE/RSJ}, 2017.

\bibitem{Starke2018-pb}
S.~Starke, N.~Hendrich, and J.~Zhang, ``Memetic evolution for generic full-body
  inverse kinematics in robotics and animation,'' \emph{IEEE Transactions on},
  2018.

\bibitem{thomas1995illusion}
F.~Thomas, O.~Johnston, and F.~Thomas, \emph{The illusion of life: Disney
  animation}.\hskip 1em plus 0.5em minus 0.4em\relax Hyperion New York, 1995.

\bibitem{shrestha2023feva}
S.~Shrestha, W.~Sentosatio, H.~Peng, C.~Fermuller, and Y.~Aloimonos, ``Feva:
  Fast event video annotation tool,'' \emph{arXiv preprint arXiv:2301.00482},
  2023.

\bibitem{pnueli1977temporal}
A.~Pnueli, ``The temporal logic of programs,'' in \emph{18th Annual Symposium
  on Foundations of Computer Science (sfcs 1977)}.\hskip 1em plus 0.5em minus
  0.4em\relax ieee, 1977, pp. 46--57.

\bibitem{kesten1998algorithmic}
Y.~Kesten, A.~Pnueli, and L.-o. Raviv, ``Algorithmic verification of linear
  temporal logic specifications,'' in \emph{International Colloquium on
  Automata, Languages, and Programming}.\hskip 1em plus 0.5em minus 0.4em\relax
  Springer, 1998, pp. 1--16.

\bibitem{konur2013survey}
S.~Konur, ``A survey on temporal logics for specifying and verifying real-time
  systems,'' \emph{Frontiers of Computer Science}, vol.~7, no.~3, pp. 370--403,
  2013.

\bibitem{finucane2010ltlmop}
C.~Finucane, G.~Jing, and H.~Kress-Gazit, ``Ltlmop: Experimenting with
  language, temporal logic and robot control,'' in \emph{2010 IEEE/RSJ
  International Conference on Intelligent Robots and Systems}.\hskip 1em plus
  0.5em minus 0.4em\relax IEEE, 2010, pp. 1988--1993.

\bibitem{guo2013revising}
M.~Guo, K.~H. Johansson, and D.~V. Dimarogonas, ``Revising motion planning
  under linear temporal logic specifications in partially known workspaces,''
  in \emph{2013 IEEE international conference on robotics and
  automation}.\hskip 1em plus 0.5em minus 0.4em\relax IEEE, 2013, pp.
  5025--5032.

\bibitem{baran2021ros}
R.~Baran, X.~Tan, P.~Varnai, P.~Yu, S.~Ahlberg, M.~Guo, W.~S. Cortez, and D.~V.
  Dimarogonas, ``A ros package for human-in-the-loop planning and control under
  linear temporal logic tasks,'' in \emph{2021 IEEE 17th International
  Conference on Automation Science and Engineering (CASE)}.\hskip 1em plus
  0.5em minus 0.4em\relax IEEE, 2021, pp. 2182--2187.

\bibitem{wang2021learning}
C.~Wang, C.~Ross, Y.-L. Kuo, B.~Katz, and A.~Barbu, ``Learning a
  natural-language to ltl executable semantic parser for grounded robotics,''
  in \emph{Conference on Robot Learning}.\hskip 1em plus 0.5em minus
  0.4em\relax PMLR, 2021, pp. 1706--1718.

\bibitem{liu2022lang2ltl}
J.~X. Liu, Z.~Yang, B.~Schornstein, S.~Liang, I.~Idrees, S.~Tellex, and
  A.~Shah, ``Lang2ltl: Translating natural language commands to temporal
  specification with large language models,'' in \emph{Workshop on Language and
  Robotics at CoRL 2022}, 2022.

\bibitem{li2021explaining}
S.~Li, M.~Feng, L.~Wang, A.~Essofi, Y.~Cao, J.~Yan, and L.~Song, ``Explaining
  point processes by learning interpretable temporal logic rules,'' in
  \emph{International Conference on Learning Representations}, 2021.

\bibitem{saveri2023towards}
G.~Saveri and L.~Bortolussi, ``Towards invertible semantic-preserving
  embeddings of logical formulae,'' \emph{arXiv preprint arXiv:2305.03143},
  2023.

\bibitem{duret2016spot}
A.~Duret-Lutz, A.~Lewkowicz, A.~Fauchille, T.~Michaud, E.~Renault, and L.~Xu,
  ``Spot 2.0—a framework for ltl and-automata manipulation,'' in
  \emph{International Symposium on Automated Technology for Verification and
  Analysis}.\hskip 1em plus 0.5em minus 0.4em\relax Springer, 2016, pp.
  122--129.

\bibitem{duret2022spot}
A.~Duret-Lutz, E.~Renault, M.~Colange, F.~Renkin, A.~Gbaguidi~Aisse,
  P.~Schlehuber-Caissier, T.~Medioni, A.~Martin, J.~Dubois, C.~Gillard
  \emph{et~al.}, ``From spot 2.0 to spot 2.10: What’s new?'' in
  \emph{International Conference on Computer Aided Verification}.\hskip 1em
  plus 0.5em minus 0.4em\relax Springer, 2022, pp. 174--187.

\bibitem{raffel2020exploring}
C.~Raffel, N.~Shazeer, A.~Roberts, K.~Lee, S.~Narang, M.~Matena, Y.~Zhou,
  W.~Li, and P.~J. Liu, ``Exploring the limits of transfer learning with a
  unified text-to-text transformer,'' \emph{The Journal of Machine Learning
  Research}, vol.~21, no.~1, pp. 5485--5551, 2020.

\bibitem{Kalegina2018-ki}
A.~Kalegina, G.~Schroeder, A.~Allchin, K.~Berlin, and M.~Cakmak,
  ``Characterizing the design space of rendered robot faces,'' in
  \emph{Proceedings of the 2018 {ACM/IEEE} International Conference on
  {Human-Robot} Interaction}, ser. HRI '18.\hskip 1em plus 0.5em minus
  0.4em\relax New York, NY, USA: Association for Computing Machinery, Feb.
  2018, pp. 96--104.

\bibitem{Takashima2008-tz}
K.~Takashima, Y.~Omori, Y.~Yoshimoto, Y.~Itoh, Y.~Kitamura, and F.~Kishino,
  ``Effects of avatar's blinking animation on person impressions,'' in
  \emph{Graphics Interface}.\hskip 1em plus 0.5em minus 0.4em\relax
  researchgate.net, 2008, pp. 169--176.

\bibitem{Trutoiu2011-se}
L.~C. Trutoiu, E.~J. Carter, I.~Matthews, and J.~K. Hodgins, ``Modeling and
  animating eye blinks,'' \emph{ACM Trans. Appl. Percept.}, vol.~8, no.~3, pp.
  1--17, Aug. 2011.

\bibitem{BaxterName_Wikipedia-qz}
{Wikipedia contributors}, ``Baxter (name),''
  \url{https://en.wikipedia.org/wiki/Baxter_(name)}, Aug. 2021, accessed:
  NA-NA-NA.

\bibitem{Fothergill2012-sb}
S.~Fothergill, H.~Mentis, P.~Kohli, and S.~Nowozin, ``Instructing people for
  training gestural interactive systems,'' in \emph{Proceedings of the {SIGCHI}
  Conference on Human Factors in Computing Systems}.\hskip 1em plus 0.5em minus
  0.4em\relax New York, NY, USA: Association for Computing Machinery, May 2012,
  pp. 1737--1746.

\bibitem{Charbonneau2011-yi}
E.~Charbonneau, A.~Miller, and J.~J. LaViola, ``Teach me to dance: exploring
  player experience and performance in full body dance games,'' in
  \emph{Proceedings of the 8th International Conference on Advances in Computer
  Entertainment Technology}, ser. ACE '11, no. Article 43.\hskip 1em plus 0.5em
  minus 0.4em\relax New York, NY, USA: Association for Computing Machinery,
  Nov. 2011, pp. 1--8.

\bibitem{Cauchard2015-vv}
J.~R. Cauchard, J.~L. E, K.~Y. Zhai, and J.~A. Landay, ``Drone \& me: an
  exploration into natural human-drone interaction,'' in \emph{Proceedings of
  the 2015 {ACM} International Joint Conference on Pervasive and Ubiquitous
  Computing}, ser. UbiComp '15.\hskip 1em plus 0.5em minus 0.4em\relax New
  York, NY, USA: Association for Computing Machinery, Sep. 2015, pp. 361--365.

\bibitem{furgale2013unified}
P.~Furgale, J.~Rehder, and R.~Siegwart, ``Unified temporal and spatial
  calibration for multi-sensor systems,'' in \emph{2013 IEEE/RSJ International
  Conference on Intelligent Robots and Systems}.\hskip 1em plus 0.5em minus
  0.4em\relax IEEE, 2013, pp. 1280--1286.

\bibitem{ffmpeg-hwaccelIntro-th}
{FFmpeg.org}, ``{FFmpeg} hardware acceleration introduction,''
  \url{https://trac.ffmpeg.org/wiki/HWAccelIntro}, 2021, accessed: 2021-9-29.

\bibitem{Liu2021-ph}
X.~Liu, H.~Shi, H.~Chen, Z.~Yu, X.~Li, and G.~Zhao, ``{iMiGUE}: An
  {Identity-Free} video dataset for {Micro-Gesture} understanding and emotion
  analysis,'' in \emph{Proceedings of the {IEEE/CVF} Conference on Computer
  Vision and Pattern Recognition}.\hskip 1em plus 0.5em minus 0.4em\relax
  openaccess.thecvf.com, 2021, pp. 10\,631--10\,642.

\bibitem{Otter-asr-vw}
Otter.ai, ``Otter: Automatic speech recognition tool,''
  \url{https://otter.ai/}, 2021, accessed: 2021-9-29.

\bibitem{Cao2019-iv}
Z.~Cao, G.~Hidalgo, T.~Simon, S.-E. Wei, and Y.~Sheikh, ``{OpenPose}: realtime
  multi-person {2D} pose estimation using part affinity fields,'' \emph{IEEE
  Trans. Pattern Anal. Mach. Intell.}, vol.~43, no.~1, pp. 172--186, 2019.

\bibitem{cohen1960coefficient}
J.~Cohen, ``A coefficient of agreement for nominal scales,'' \emph{Educational
  and psychological measurement}, vol.~20, no.~1, pp. 37--46, 1960.

\bibitem{yang2015robot}
Y.~Yang, Y.~Li, C.~Fermuller, and Y.~Aloimonos, ``Robot learning manipulation
  action plans by" watching" unconstrained videos from the world wide web,'' in
  \emph{Proceedings of the AAAI Conference on Artificial Intelligence},
  vol.~29, no.~1, 2015.

\bibitem{Abner2015-ot}
N.~Abner, K.~Cooperrider, and S.~Goldin-Meadow,
  ``\BIBforeignlanguage{en}{Gesture for linguists: A handy primer},''
  \emph{\BIBforeignlanguage{en}{Lang. Linguist. Compass}}, vol.~9, no.~11, pp.
  437--451, Nov. 2015.

\bibitem{lyons1977semantics}
\BIBentryALTinterwordspacing
J.~Lyons, \emph{Semantics: Volume 1}, ser. ACLS Humanities E-Book.\hskip 1em
  plus 0.5em minus 0.4em\relax Cambridge University Press, 1977. [Online].
  Available: \url{https://books.google.com/books?id=jQA6wVLCINUC}
\BIBentrySTDinterwordspacing

\end{thebibliography}

\end{document}